\documentclass[sigconf]{acmart}

\usepackage{fancyhdr}
\usepackage{amsmath}
\usepackage{booktabs}
\usepackage{bbding}
\usepackage{enumitem}
\usepackage{multirow}
\usepackage{amsthm}
\usepackage{mathrsfs}
\usepackage{diagbox}
\usepackage{bm}
\usepackage{newfloat}
\usepackage{xspace}
\usepackage{comment}
\usepackage{bbm}
\usepackage{listings}
\usepackage{makecell}
\usepackage{graphicx}

\DeclareMathOperator*{\argmax}{arg\,max}

\newcommand{\system}{BMB\xspace}
\newcommand{\ie}{\textit{i}.\textit{e}.} 
\newcommand{\fakeparagraph}[1]{\vspace{3mm}\noindent\textbf{#1}}

\usepackage[capitalize]{cleveref}
\crefname{section}{Sec.}{Secs.}
\Crefname{section}{Section}{Sections}
\Crefname{table}{Table}{Tables}
\crefname{table}{Tab.}{Tabs.}

\begin{document}

\title{BMB: Balanced Memory Bank for Imbalanced \\ Semi-supervised Learning}

\author{Wujian Peng}
\affiliation{%
  \institution{Fudan University}
  \country{}}

\author{Zejia Weng}
\affiliation{%
  \institution{Fudan University}
  \country{}}

\author{Hengduo Li}
\affiliation{%
  \institution{University of Maryland}
  \country{}}
\email{}

\author{Zuxuan Wu}
\affiliation{%
  \institution{Fudan University}
  \country{}}
\authornote{Corresponding author.}

\renewcommand{\shorttitle}{BMB: Balanced Memory Bank for Imbalanced Semi-supervised Learning}

\renewcommand\footnotetextcopyrightpermission[1]{}
\settopmatter{printacmref=false}

\begin{abstract}
Exploring a substantial amount of unlabeled data, semi-supervised learning~(SSL) boosts the recognition performance when only a limited number of labels are provided. However, traditional methods assume that the data distribution is class-balanced, which is difficult to achieve in reality due to the long-tailed nature of real-world data.
While the data imbalance problem has been extensively studied in 
supervised learning (SL) paradigms, directly transferring existing approaches to SSL is nontrivial, as prior knowledge about data distribution remains unknown in SSL. In light of this, we propose Balanced Memory Bank~(\system), a semi-supervised framework for long-tailed recognition. The core of \system is an online-updated memory bank that caches historical features with their corresponding pseudo labels, and the memory is also  carefully maintained to ensure the data therein are class-rebalanced. Additionally, an adaptive weighting module is introduced to work jointly with the memory bank so as to further re-calibrate the biased training process. We conduct  experiments on multiple datasets and demonstrate, among other things, that \system surpasses state-of-the-art approaches by clear margins, for example 8.2\(\%\) on the 1\(\%\) labeled subset of ImageNet127 (with a resolution of \(64\times64\)) and 4.3\(\%\) on the 50\(\%\) labeled subset of ImageNet-LT.
\end{abstract}




\maketitle

\section{Introduction}

\begin{figure}
    \centering
    \includegraphics[width=1.\linewidth]{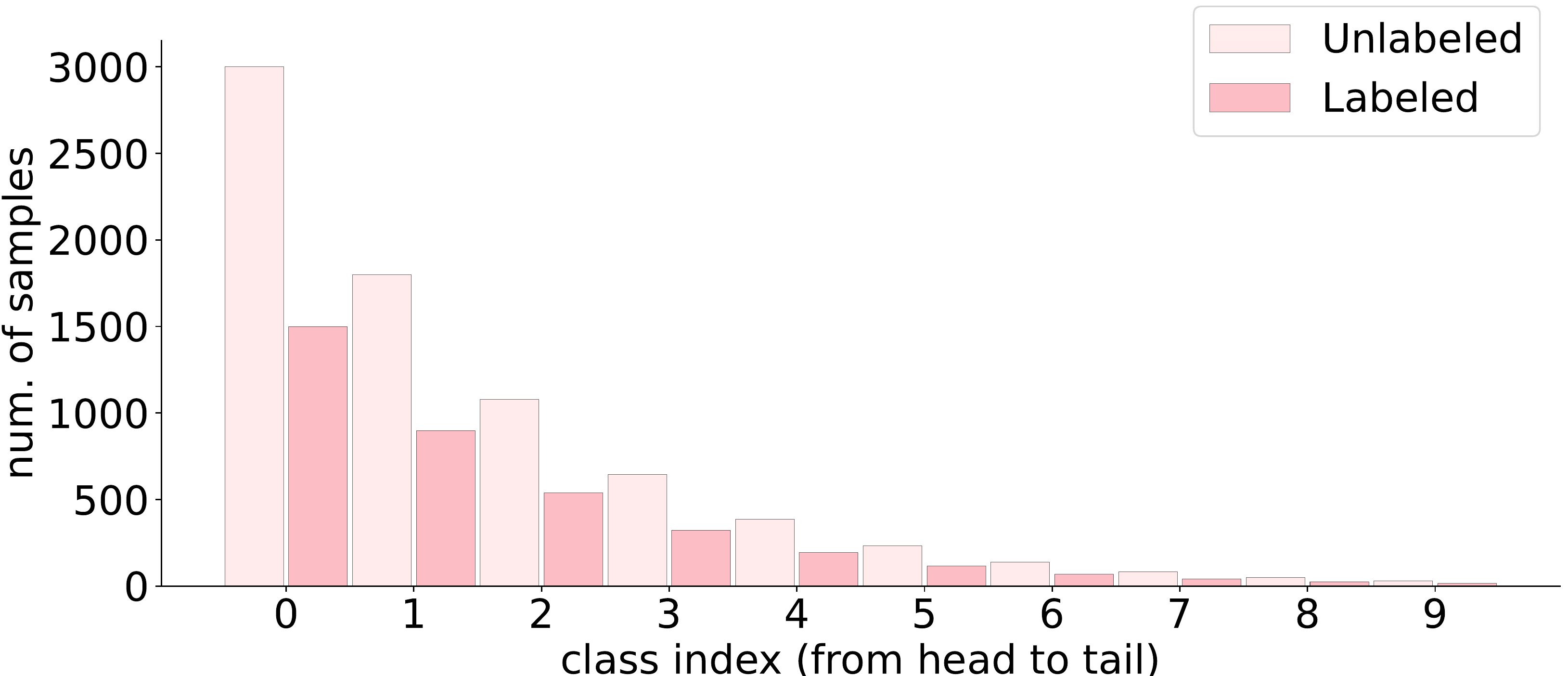}
    {(a) Long-tailed distribution of CIFAR10-LT.}
    \includegraphics[width=1.\linewidth]{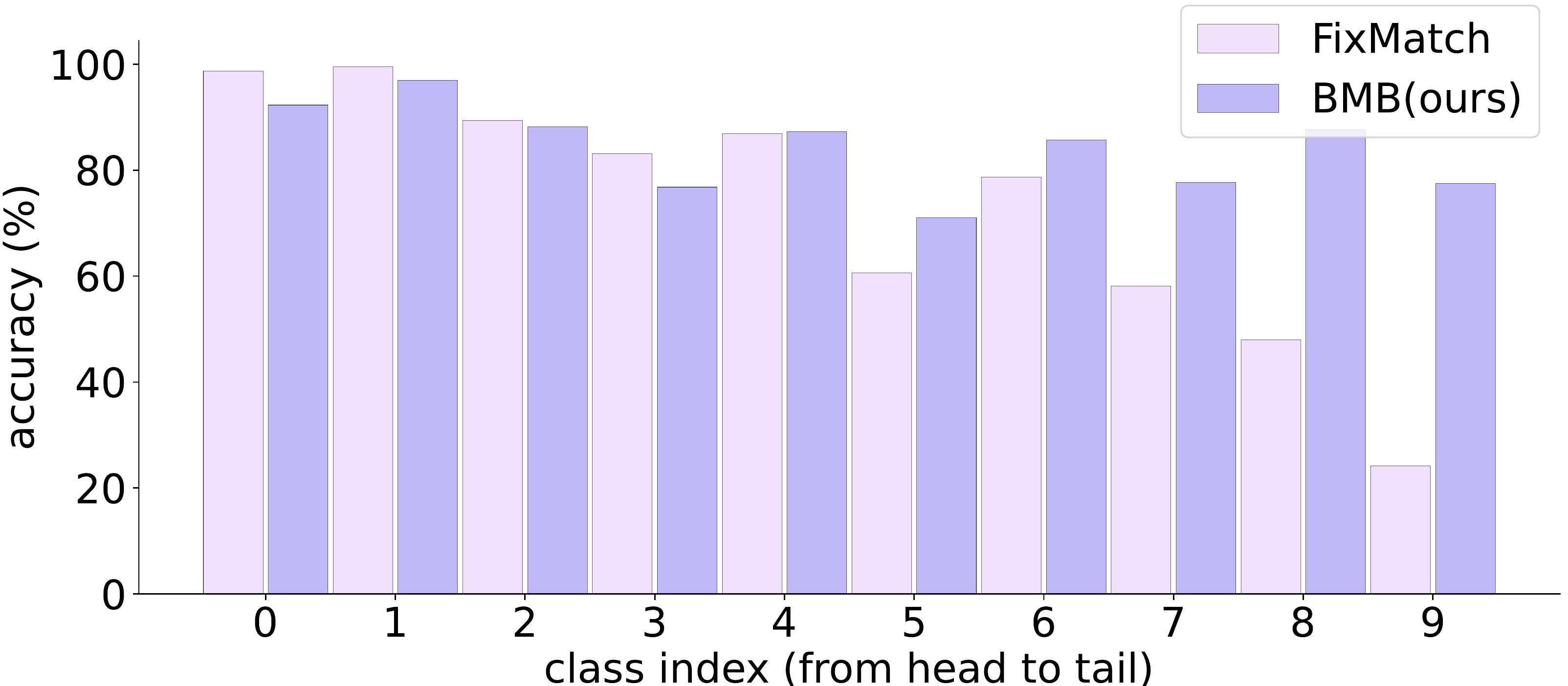}
    {(b) Accuracy of each category on CIFAR10-LT.}
    \caption{(a) Both labeled and unlabeled data follow a long-tailed distribution in imbalanced SSL. Classes with more samples are referred to as majority classes, while those with fewer samples are referred to as minority classes. (b) Conventional SSL algorithms perform poorly on minority classes, with the help of BMB, the model exhibits significant accuracy increase for the minority classes, while maintaining comparable accuracy for the majority classes.}
    \label{fig:teaser}
    \vspace{-1em}
\end{figure}

Semi-supervised learning (SSL) aims to learn from large amounts of unlabeled data together with a limited number of labeled data so as to mitigate the need for costly large-scale manual annotations. Although extensive studies have shown that deep neural networks can achieve high accuracy even with limited samples when trained in the semi-supervised manner~\cite{berthelot2019mixmatch, sohn2020fixmatch,uda,mean-teahcer,noisystudent}, the majority of existing approaches assume that the distribution of labeled and unlabeled data are class-balanced. This is in stark contrast to realistic scenarios where data are oftentimes long-tailed, \ie , the majority of samples belong to a few dominant classes while the remaining classes have far fewer samples, as illustrated in Figure~\ref{fig:teaser}(a). 

The long-tailed nature of classes makes it particularly challenging for SSL compared to conventional supervised training pipelines. This results from the fact that the mainstream of SSL relies on pseudo labels produced by teacher networks~\cite{sohn2020fixmatch,mean-teahcer,noisystudent}, which is trained with a handful of labeled samples that are drawn from a skewed class distribution. As a result, these generated pseudo labels are biased towards the majority classes and thus the class imbalance is further amplified, resulting in deteriorated performance particularly on minority classes, as shown in Figure~\ref{fig:teaser}(b). 

One popular strategy to mitigate the class imbalance problem in long-tailed supervised learning (SL) is data re-sampling, which balances the training data by under-sampling the majority classes and over-sampling the minority classes. While seemingly promising, generalizing the re-sampling method from SL to SSL is non-trivial since the method requires knowledge about the labels and class distribution of training data, which are missing in SSL that mainly learns from \emph{unlabeled} data. As a result, existing re-sampling approaches for SSL still produce relatively unsatisfactory performance~\cite{He2021RethinkingRI,wei2021crest,lee2021abc}. It is clear that the SSL performance would be further improved with better-tailored re-sampling strategies that can bridge the gap mentioned above between SL and SSL. Motivated by this, we attempt to address the challenges encountered on re-sampling in SSL and demonstrate that re-sampling can also achieve good results in class-imbalanced SSL.

With this in mind, we introduce \textbf{B}alanced \textbf{M}emory \textbf{B}ank~(\system), a semi-supervised framework for long-tailed classification. \system contains a balanced feature memory bank and an adaptive weighting module, cooperating with each other to re-calibrate the training process. In particular, the balanced feature memory bank stores historical features of unlabeled samples with their corresponding pseudo labels that are updated online. During training, a certain number of pseudo-annotated features are selected from the memory bank to supplement features in the current batch, and features of the minority classes are more likely to be chosen to enhance the classifier's capacity for classifying the tail categories. It is worth noting that when inserting features into the memory bank, we update the memory with only a subset of samples to keep the memory bank class-rebalanced instead of storing all samples from the current batch, ensuring the model to learn from a diverse set of data. 
In addition, the adaptive weighting module aims to address the class imbalance issue in SSL by assigning higher weights to the losses of samples from minority classes and lower weights to those from majority ones, which enables the model to learn a more balanced classifier.

We conduct experiments on the commonly-studied datasets CIFAR10-LT~\cite{cifar} and CIFAR100-LT~\cite{cifar} and show that \system achieves  better performance than previous state-of-the-arts. As demonstrated in Figure~\ref{fig:teaser}(b), with the help of \system, the accuracy of minority classes exhibits significant boost compared to the baseline model~\cite{sohn2020fixmatch}. Furthermore, we also conduct experiments on larger-scale datasets, ImageNet127~\cite{imagenet127} and ImageNet-LT~\cite{imagenet-lt}, which are more realistic and challenging. \system outperforms state-of-the-art approaches with clear margins, highlighting its effectiveness in more practical settings. 
Specifically, compared to the previous state-of-the-arts, \system achieved improvements of 8.2\(\%\) on the 1\(\%\) labeled subset of ImageNet127 (with a resolution of 64\(\times\)64) and 4.3\(\%\) on the 50\(\%\) labeled subset of ImageNet-LT.
It is worth pointing out that we are the first to evaluate class-imbalanced semi-supervised algorithms on ImageNet-LT~\cite{imagenet-lt}, which is a more challenging benchmark with up to 1,000 categories, making it more difficult to handle the bias in pseudo labels. Besides, the imbalance in ImageNet-LT is more severe (the rarest class only contains 5 samples) which will be even fewer in semi-supervised setting. This makes the modeling for the minority class more difficult. We believe the class-imbalanced SSL should focus more on such realistic and challenging benchmarks to drive 
further progress.

The main contributions of this paper are summarized as follows:
\begin{itemize}
  \item We present \system, a novel semi-supervised learning framework for class-imbalanced classification. It comprises a balanced memory bank and an adaptive weighting module, which work collaboratively to rebalance the learning process in class-imbalanced SSL.
  \item We conduct extensive experiments on various datasets to verify the effectiveness of \system, and achieve state-of-the-art results on several benchmarks. Notably, we pioneered the experimentation with ImageNet-LT, which provides a more challenging and realistic benchmark for future works.
\end{itemize}

\section{RELATED WORK}
\subsection{Semi-supervised Learning}
To mitigate the expensive data annotation cost in SL, a range of approaches aim to learn from unlabeled data, in order to enhance the performance on limited labeled data. One widely used approach is consistency regularization~\cite{temporal-ensembling,vat,mean-teahcer} that enforces consistent predictions for similar inputs, serving as a regularization term during training. Pseudo labeling~\cite{pseudo-label,noisystudent} is another line of research that  assigns pseudo labels to unlabeled data based on the predictions of a teacher model. When pseudo labels are assigned by the model itself, this is generally known as self-training~\cite{uda,noisystudent}. FixMatch~\cite{sohn2020fixmatch} builds upon both consistency regularization and pseudo labeling, and presents state-of-the-art performance on class-balanced datasets, but produces limited results when the data distribution is imbalanced. Our approach differs from the standard SSL method that we wish to explicitly build a class-balanced classifier by a balanced memory bank that alleviates the difficulties of SSL under long-tailed datasets.

\begin{figure*}[h]
  \centering
  \includegraphics[width=0.8\linewidth]{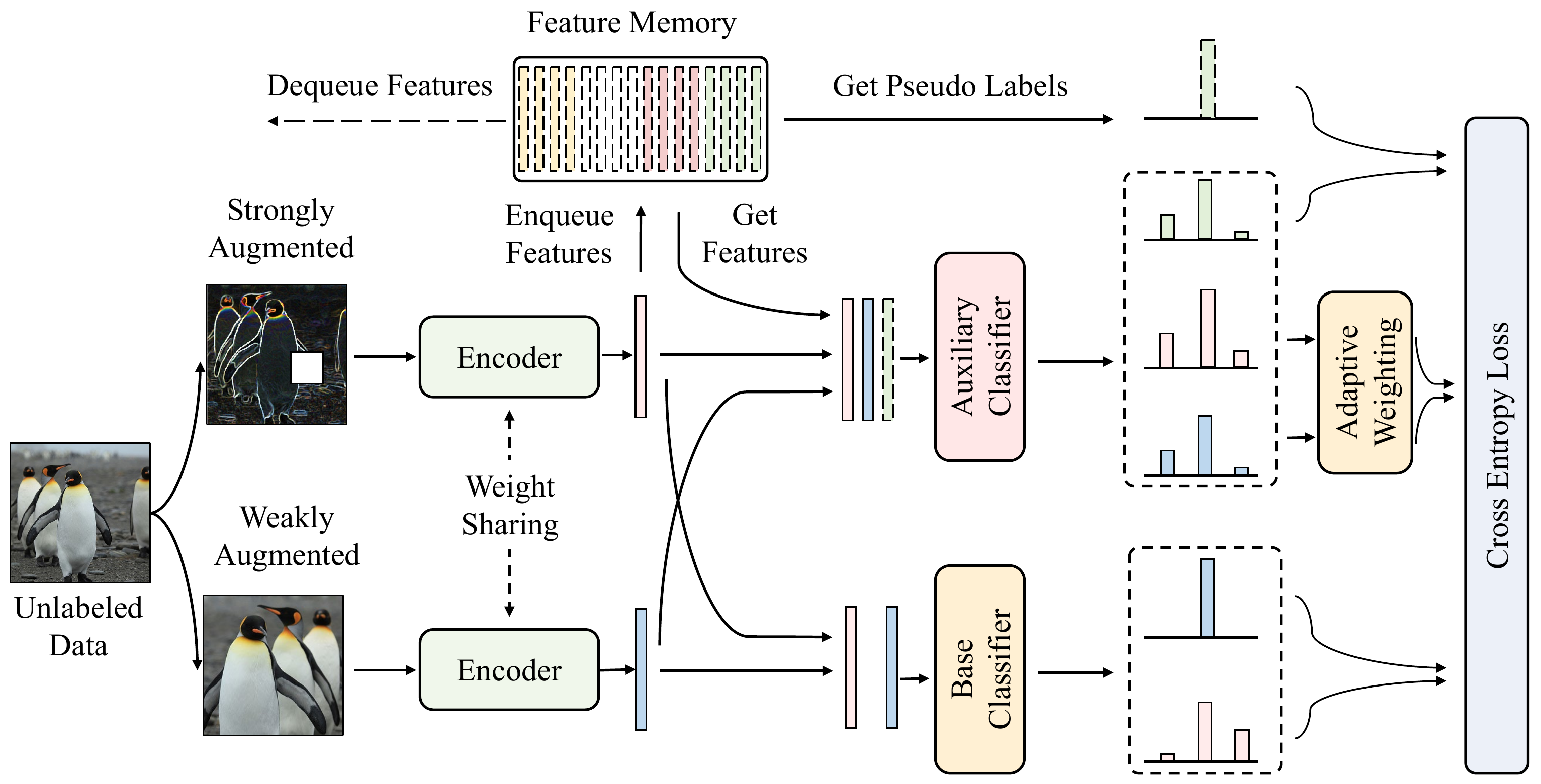}
  \caption{The overall framework of BMB, which consists of a shared encoder and two separate classifiers, \ie ~an auxiliary and a base classifier, respectively. The auxiliary classifier is trained carefully to avoid being biased towards the majority classes. The base classifier is responsible for facilitating the training of the encoder to extract better features. During inference, only the balanced auxiliary classifier is used while the base classifier is discarded.}
  \label{fig:framework}
\end{figure*}

\subsection{Class-imbalanced Supervised Learning}
Real-world data usually exhibit a long-tailed distribution, with a significant variance in the number of samples across different categories. To improve the performance of tail classes, re-weighting methods~\cite{lin2017focal,cui2019class,cao2019learning} assign a higher loss weight for the minority classes and a lower one for the majority classes, forcing the model to pay more attention to the minorities. Re-sampling approaches~\cite{chawla2002smote,he2009learning,buda2018systematic} attempt to achieve re-balancing at the sampling level, \ie, minority-classes are over-sampled or majority-classes are under-sampled. However, this usually leads to overfitting or information loss~\cite{chawla2002smote, cui2019class}. In addition, two-stage training approaches~\cite{kang2019decoupling,zhou2020bbn} decouple the learning of representations and the classifiers. The feature extractor is obtained in the first stage, and a balanced classifier is trained in the second stage with the extractor fixed. More recently, logits compensation~\cite{LA,Tan2020EqualizationLF} and contrastive-based methods~\cite{li2022targeted,wang2021contrastive,zhu2022balanced} also  show promising performance. These methods resort to the known data distributions to achieve re-balancing among different classes. However, this information is unknown for unlabeled dataset in semi-supervised scenario.

\subsection{Class-imbalanced Semi-supervised Learning}
There is a growing interest in the class-imbalanced problem for SSL. However, it is extremely challenging to deal with the class-imbalanced data in SSL due to the unknown data distribution and the unreliable pseudo labels provided by a biased teacher model. DARP~\cite{DARP} formulates a convex optimization to refine inaccurate pseudo labels. CReST~\cite{wei2021crest} selectively chooses unlabeled data to complement the labeled set, and the minority classes are selected with a higher frequency. DASO~\cite{DASO} introduces a semantic-aware feature classifier to refine pseudo labels. CoSSL~\cite{fan2022cossl} disentangles the training of the feature extractor and the classifier head, and introduces interaction modules to couple them closely. 
Unlike these approaches, we address class-imbalance through re-sampling with the help of a memory bank to update pseudo labels in an online manner, while estimating the distribution of the unlabeled data through a simple yet effective approach. This allows for an end-to-end training pipeline in a single stage.

\section{Preliminary: A Semi-supervised Framework}
\subsection{Notation}
We assume a semi-supervised dataset contains $N$ labeled samples and $M$ unlabeled samples and refer to the labeled set as \(\mathcal{X}=\{(x_i,y_i)\}_{i=1}^{N}\) and the unlabeled set as \(\mathcal{U}=\{u_j\}_{j=1}^{M}\), respectively. We use the index \(i\) for labeled data, the index \(j\) for unlabeled data and index \(k\) for the label space. The number of training samples in the \(k\)-\(th\) class is denoted as \(N_k\) and \(M_k\) for the labeled and unlabeled set, respectively, \ie, \(\sum_{k=1}^{K}{N_k}=N\) and \(\sum_{k=1}^{K}{M_k}=M\). Without loss of generality, we let \(N_1\ge N_2 \ge \cdots \ge N_K\) for simplicity. We use $f (x; \theta)$ to represent the mapping function of the model, $\alpha$ and $\mathcal{A}$ to represent the weak augmentation and the strong augmentation respectively. We use $\gamma_l=\frac{N_1}{N_K}$ and $\gamma_u=\frac{M_1}{M_K}$ to reflect the imbalance ratios for the labeled and unlabeled datasets respectively.
\subsection{FixMatch}
\label{sec:fixmatch}
FixMatch~\cite{sohn2020fixmatch} is one of the most popular SSL algorithms that enables deep neural networks to effectively learn from unlabeled data. 
A labeled example \(x_i\)  is first transformed to its weakly augmented version \(\alpha(x_i)\) and then taken as input by the model $f$. The supervised loss during training is calculated following \cref{eq:ls_backbone}:
\begin{equation}
    \mathcal{L}_s=\frac{1}{B}\sum_{i=1}^B{\mathbf{H}(y_i,f(\alpha(x_i)))}
    \label{eq:ls_backbone}
\end{equation}
where \(B\) refers to the batch size, \(y_i\) is the label of \(x_i\), and \(\mathbf{H}(\cdot,\cdot)\) denotes the standard cross-entropy loss.

Given an unlabeled sample $u_j$, two different views $\mathcal{A}(u_j)$ and $\alpha(u_j)$ are obtained by applying the strong augmentation $\mathcal{A}$ and the weak augmentation $\alpha$ to the sample. The predicted probability vector on $u_j$ is denoted as $\mathbf{q}_j=f(\alpha(u_j))$, which is then converted into a pseudo categorical label: \(\hat{q}_j=\argmax(\mathbf{q}_j)\) as the supervisory signal for the unlabeled sample. Finally, a cross-entropy loss is computed on the prediction of the strongly augmented view \(\mathcal{A}(u_j)\):
\begin{equation}
    \mathcal{L}_u=\frac{1}{B}\sum_{j=1}^B\mathbb{I}(max(\mathbf{q}_j)\ge\tau){\mathbf{H}(\hat{q}_j,f(\mathcal{A}(u_j)))}
    \label{eq:lu_backbone}
\end{equation}
where \(\tau\) denotes the threshold for filtering out those low-confidence and  potentially noisy pseudo labels, and $\mathbb{I}$ is the indicator function.

The total training loss is the sum of both supervised and unsupervised~\footnote{Here we slightly abuse the term ``unsupervised loss'' as the loss on unlabeled samples.} losses: \(\mathcal{L}=\mathcal{L}_s+\lambda_u\mathcal{L}_u\), where \(\lambda_u\) is a hyperparameter controlling the weight of the unsupervised loss.

\section{Our Approach}

Our goal is to develop a SSL framework for long-tailed classification with minimal surgery to the standard SSL training process, and effectively alleviating the issue of class imbalance. To this end, we present \system, an effective framework with an online-updated memory bank storing class-rebalanced features and their corresponding pseudo labels. The carefully designed memory bank serves as an additional source of training data for the classifier to cope with imbalanced class distributions. To further emphasize the minority classes during training, we also utilize a re-weighting strategy to adaptively assign weights to the loss terms for different samples. This ensures a more stable memory updating process especially during the initial stage of training.

\subsection{Overall Framework}
Previous studies~\cite{kang2019decoupling,fan2022cossl} have shown that imbalanced training data have little impact on encoders (\ie, feature extractors), and the bias towards majority classes mainly occurs in classifiers. To balance the classifier, there are studies~\cite{zhou2020bbn, lee2021abc} introducing an additional branch to assist the learning process. Inspired by this, we build our \system on top of the conventional SSL framework by equipping it with an extra classifier, and ensure it to be class-rebalanced through carefully designed techniques. As depicted in Figure~\ref{fig:framework}, \system comprises a shared feature encoder and two distinct classifiers. 

More specifically, each classifier performs its own role in the whole framework, and with one referred to as the base classifier and another one as the auxiliary classifier, respectively. The base classifier aims to help the encoder to extract better features, and its training follows the traditional SSL methods without any additional re-balancing operation. In contrast, the auxiliary classifier is responsible for making a reliable prediction without biasing towards the majority classes. To make the auxiliary classifier more balanced, we introduce a memory bank that caches historical features to provide more balanced training data, and a loss re-weighting strategy is utilized to ensure the memory bank being well-initialized and maintained. 

The training process of \system is end-to-end, and all the components are jointly trained. During inference, the base classifier is discarded, and the output of the auxiliary classifier is used as the final prediction. 

\subsection{Balanced Feature Memory Bank}
\label{sec:memory-bank}
\begin{figure}[t]
  \centering
  \includegraphics[width=\linewidth]{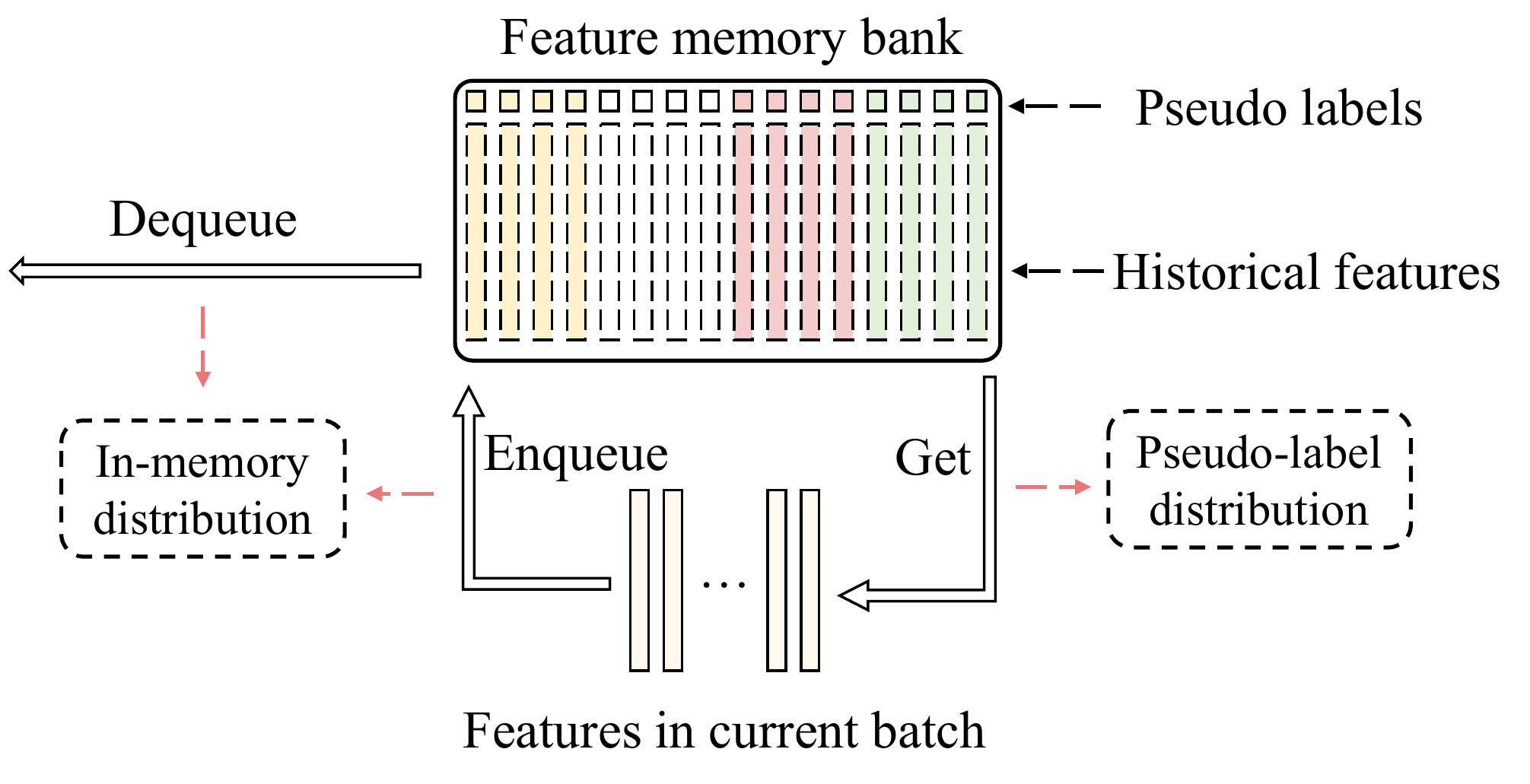}
  \caption{The maintenance mechanism of the feature memory bank. The \(enqueue\) and \(dequeue\) operation is based on the current class distribution in the memory bank, and their goal is to make the memory class-balanced. The \(get\) operation samples features from the memory according to the estimated unlabeled data distribution, and the minority-class features are selected with a higher probability to complement features in the current batch.}
  \label{fig:memory_bank}
\end{figure}
We construct a memory bank structure with a fixed storage size to cache historical features and their corresponding pseudo labels. The memory bank consists of three key operations: \textit{enqueue}, \textit{dequeue}, and \textit{get}, which are crucial for maintaining and utilizing the memory bank effectively. The \textit{enqueue} operation adds features to the memory bank, while the \textit{dequeue} operation eliminates unnecessary features when the memory bank reaches its maximum capacity. During training with the memory bank, the \textit{get} operation retrieves data from the memory based on a predefined strategy to supplement the features in the current batch. Figure~\ref{fig:memory_bank} illustrates the memory bank structure and these operations.

\fakeparagraph{Enqueue and Dequeue.}
The basic intuition of maintaining the memory bank is to keep it category balanced. Specifically, let \(C_k\)  denote the count of features in the memory belonging to the \(k\)-\(th\) class, we aim to ensure the in-memory distribution \((C_1,\cdots,C_K)\) is as uniform as possible. As such, we carefully design the updating strategy accomplished by the \textit{enqueue} and \textit{dequeue} operations.

For each training step, the \textit{enqueue} operation adds the most recent features to the memory with a varying probability. Specifically, if a feature has been confidently pseudo-annotated as belonging to the $k$-th class category, it is put into the memory with a probability based on the number of features for the $k$-th class in the bank:
\begin{equation}
    P_k^{in}=\frac{1}{(C_k)^{\beta}}
    \label{eq:put_prob}
\end{equation}
where $\beta$ is a hyperparameter larger than 0. With \cref{eq:put_prob}, features from categories that are seldom seen are more likely to be put into the memory bank. 

When the memory bank reaches its maximum capacity, incoming features and their pseudo labels will need to replace existing ones in the memory bank. In this case, we use the \textit{dequeue} operation to discard a certain number of features and their pseudo labels. To maintain a class-balanced memory bank, we remove the majority features with a higher probability, while the minority features are removed with a lower probability calculated as follows:
\begin{equation}
    P_k^{out}=1-\frac{1}{(C_k)^{\beta}}
    \label{eq:remove_prob}
\end{equation}
where \(\beta\) \(>0\) is a coefficient that controls the balance level of the memory, and a larger value makes the memory more balanced.

\fakeparagraph{Get.}
After obtaining a class-rebalanced memory bank, we design an algorithm to perform re-sampling at the feature level via the \textit{get} operation, aiming to balance the auxiliary classifier. We employed reversed sampling based on the distribution of training samples to compensate the imbalance in the current batch and thus eliminating the bias effects of the long-tail phenomenon. Specifically, features that belong to the $k$-th class will be sampled with the probability described in:

\begin{equation}
    P_k^{get}=\frac{1}{({M}_k)^{\lambda}}
    \label{eq:get_prob}
\end{equation}
where \({M}_k\) refers to the number of unlabeled training data belonging to class $k$, and \(\lambda\) controls the level of reversed sampling. With a larger \(\lambda\), the minority classes will be over-sampled, which can compensate for the imbalanced data in current batch. The sampled features and the corresponding pseudo labels are used in the training process of the auxiliary classifier, with the corresponding loss term denoted as \(\mathcal{L}_{mem}\).

\fakeparagraph{Unlabeled data distribution estimation.}
The re-sampling operation relies on the distribution information~(\ie, the number of samples contained in each category) of the unlabeled data, which is not available in SSL. Therefore, it is necessary to estimate it appropriately. A straightforward approach is to use the labeled data distribution as a proxy, assuming that the training data are sampled from the same distribution, but this assumption may not hold when the distributions do not match. For a more accurate estimation, we use the number of accumulated pseudo labels to substitute $M_k$ with $\Tilde{M}_K$ as in~\cref{eq:distribution_estimate}:

\begin{equation}
    \Tilde{M}_k = \sum_{j=1}^{|\mathcal{P}|}\mathbbm{1}(p_j=k)
    \label{eq:distribution_estimate}
\end{equation}
where \(\mathcal{P}\) denotes all the pseudo labels of the unlabeled dataset,  \(p_j\) is the $j$-th pseudo label and $\mathbbm{1}$ is the indicator function. In this way, we can obtain the estimated distribution $(\Tilde{M}_1,\Tilde{M}_2,\cdots,\Tilde{M}_k)$ of the unlabeled dataset.

\subsection{Adaptive Loss Re-weighting}
\label{sec:ada-weight}

The class-rebalanced memory bank enables online re-sampling to alleviate the class imbalance issue. However, solely relying on the memory bank can be problematic since the pseudo labels may exhibit bias towards the majority classes in the early stage of training. This can lead to reduced effectiveness of the memory bank since the in-memory samples belonging to the minority classes is scarce and the pseudo labels are unreliable. Furthermore, enriching the batch with features from the memory bank itself may not be enough for perfectly balancing the majority and minority classes since the number of samples for each class within the batch is an integer, making the re-calibration during training ``discrete'' as opposed to a continuous process with more controllable variance. To this end, we propose an adaptive weighting method that not only ensures the memory bank being well-initialized and maintained, but also enables a flexible and continuous calibration with controllable variance that further mitigates the class imbalance.

Formally, for each sample \(x_i\) from the labeled set, an adaptive loss weight \(W(x_i)\) is generated to re-weight the loss for the auxiliary classifier $f_a$ as below:
\begin{equation}
    W(x_i)=\left(\frac{N_K}{N_{y_i}}\right)^\alpha
    \label{eq:adaw-labeled}
\end{equation}
where \(N_K\) is the number of samples from the class with the least samples, and \(N_{y_i}\) is the number of samples from class \(y_i\). The weight is inversely proportional to the number of samples in class \(y_i\) and the hyper-parameter \(\alpha\) controls the variance of weights, where a larger value will lead to more diverse weights across different classes. The adaptive weights are then injected into the original supervised loss as follows:
\begin{equation}
    \mathcal{L}_s^a=\frac{1}{B}\sum_{i=1}^{B}W(x_i)\mathbf{H}(y_i,f_a(\alpha(x_i)))
    \label{eq:loss-labeled}
\end{equation}

For the unlabeled sample \(u_j\), the adaptive weight is computed in a similar way, except that we replace the number of samples in ~\cref{eq:adaw-labeled} with the estimated distribution $\Tilde{M}$:
\begin{equation}
    W(u_j)=\left(\frac{\Tilde{M}_K}{\Tilde{M}_{\hat{q}_j}}\right)^\alpha
    \label{eq:adaw-unlabeled}
\end{equation}
where \(\hat{q_j}=\argmax(f_a(\alpha(u_j)))\) is the predicted pseudo label of $u_j$. The unsupervised loss of the auxiliary classifier then becomes:
\begin{equation}
    \mathcal{L}_u^a=\frac{1}{B}\sum_{j=1}^{B}W(u_j)\mathbb{I}(max(\mathbf{q}_j)\ge\tau)\mathbf{H}(\hat{q_j},f_a(\mathcal{A}(u_j)))
    \label{eq:loss-unlabeled}
\end{equation}

\subsection{Training and Inference}
\system is an end-to-end trainable framework where all modules are trained collaboratively. The total loss, defined in ~\cref{eq:loss-total}, consists two parts: one for the base classifier and the other for the auxiliary classifier.

\begin{equation}
    \mathcal{L}_{total}=\mathcal{L}_{base}+\mathcal{L}_{aux}
    \label{eq:loss-total}
\end{equation}

The loss of the base classifier, denoted as $\mathcal{L}_{base}=\mathcal{L}_s^b+\lambda_u\mathcal{L}_u^b$, is simply the weighted sum of the original supervised and unsupervised losses described in Sec.~\ref{sec:fixmatch}, the superscript $b$ here is used to distinguish from 
the auxiliary classifier. As for the auxiliary classifier, the loss can be expressed as $\mathcal{L}_{aux}=\mathcal{L}_s^a+\lambda_u\mathcal{L}_u^a+\lambda_m\mathcal{L}_{mem}$. This is also a summation over the supervised and unsupervised losses, however, the weights are adaptively adjusted as described in Sec.~\ref{sec:ada-weight}. In addition, an extra term $\mathcal{L}_{mem}$ is included to utilize the training samples selected from the class-rebalanced memory bank. There are two hyperparameters $\lambda_u$ and $\lambda_m$ used to control the weight of different part.

Due to the absence of re-balancing adjustment for the base classifier, it is expected to be biased. Therefore, during inference, we ignore it and rely solely on the prediction from the auxiliary classifier, which is considered to be more class-balanced. However, this dose not imply that the base classifier is useless. As will be shown in~\cref{sec:ablate_classfiers}, the base classifier helps extracting better features, which is crucial for the auxiliary classifier's training.

\section{Experiments}
This section describes our experimental evaluation, where we compare \system with state-of-the-art methods and  conduct ablation studies to validate the effectiveness of each design choice in \system.

\subsection{Datasets}
To validate the effectiveness of \system, we perform experiments on several datasets, including ImageNet-LT~\cite{imagenet-lt}, ImageNet127~\cite{imagenet127} and the long-tailed version of CIFAR~\cite{cifar}.

\fakeparagraph{ImageNet-LT.}
ImageNet-LT~\cite{imagenet-lt} is constructed by sampling a subset from the orignial ImageNet~\cite{Russakovsky2014ImageNetLS} dataset following the Pareto distribution with power \(\alpha\)=6. It contains 115.8K images across 1,000 categories, and the distribution is extremely imbalanced. The most frequent class has 1280 samples, while the least frequent class only has 5 samples. 
To create a semi-supervised version of this dataset, we randomly sample 20$\%$ and 50$\%$ of the training data to form the labeled set, while all remaining training data is used as the unlabeled set, with their labels ignored. Due to the challenging nature of this dataset, previous SSL algorithms have not been evaluated on it. Nevertheless, we believe that testing on such more realistic datasets is crucial.

\fakeparagraph{ImageNet127.}
ImageNet127~\cite{imagenet127} is a large-scale dataset, which groups the 1,000 categories of ImageNet~\cite{Russakovsky2014ImageNetLS} into 127 classes based on their hierarchical structure in WordNet. It is naturally long-tailed with an imbalance \(\gamma\approx256\). The most majority class contains 277,601 images, while the most minority class only has 969 images. Following~\cite{wei2021crest,fan2022cossl}, we randomly select 1$\%$ and 10$\%$ of its training sample as the labeled set, with the remaining training samples treated as the unlabeled set. The test set is also imbalanced due to the category grouping, and we keep it untouched while reporting the averaged class recall as an evaluation metric.

\fakeparagraph{CIFAR-LT.}
The original CIFAR dataset is class balanced, to achieve the predefined imbalance ratios \(\gamma\), we follow common practice~\cite{cui2019class,fan2022cossl} by randomly selecting samples for each class from the original balanced dataset~\cite{cifar}. Specifically, we select \(N_k=N_1\cdot{\gamma}^{\mu_k}\) labeled samples and \(M_k=M_1\cdot{\gamma}^{\mu_k}\) unlabeled samples for the \(k\)-\(th\) class, where \(\mu_k=-\frac{k-1}{K-1}\). For CIFAR10 we set \(N_1\)=1500, \(M_1\)=3000, and for CIFAR100, we set \(N_1\)=150 and \(M_1\)=300. The test set remains untouched and balanced.

\subsection{Implementation Details}
\begin{table}
\centering
\resizebox{0.95\columnwidth}{!}{
\renewcommand\arraystretch{1.2}
\begin{tabular}{lllll}
\toprule
& overall & many-shot & medium-shot & few-shot    \\
&  & \multicolumn{1}{c}{> 20} & \multicolumn{1}{c}{$\leq$20~\& >4} & \multicolumn{1}{c}{$\leq$4} \\ \cmidrule(lr){2-5} 
Vanilla~\cite{wider-resnet}          &  12.5  &  24.1  &  5.8   &  1.0 \\
FixMatch~\cite{sohn2020fixmatch}     &  16.5  &  32.4  &  7.2   &  1.2 \\
CReST+~\cite{wei2021crest}           &  18.0  &  33.3  &  9.4   &  1.6 \\
CoSSL~\cite{fan2022cossl}            &  19.1  &  34.5  &  11.0  &  1.9  \\
DARP~\cite{DARP}                     &  23.0  &  40.7  &  13.8  &  2.7 \\
BMB~(ours)                           &  \textbf{25.8}\small{~$\uparrow$2.8}  &  \textbf{41.6}\small{~$\uparrow$0.9}  &  \textbf{18.2}\small{~$\uparrow$4.4}  &  \textbf{5.7}\small{~$\uparrow$3.0} \\ \bottomrule 
\end{tabular}
} \\
\vspace{0.1cm}
{(a) ImageNet-LT 20$\%$ labeled subset} \\
\vspace{0.2cm}
\resizebox{0.95\columnwidth}{!}{
\renewcommand\arraystretch{1.2}
\begin{tabular}{lllll}
\toprule
& overall & many-shot & medium-shot & few-shot    \\  
&  & \multicolumn{1}{c}{> 50} & \multicolumn{1}{c}{$\leq$50~\& >10} & \multicolumn{1}{c}{$\leq$10} \\ \cmidrule(lr){2-5}
Vanilla~\cite{wider-resnet}          &  20.9  &  36.3  &  13.5   &  2.6 \\
FixMatch~\cite{sohn2020fixmatch}     &  25.2  &  44.2  &  15.9   &  3.0 \\
CReST+~\cite{wei2021crest}           &  27.3  &  45.6  &  18.9   &  5.1 \\
CoSSL~\cite{fan2022cossl}            &  28.6  &  46.9  &  20.6   &  4.7 \\
DARP~\cite{DARP}                     &  30.9  &  50.3  &  22.2   &  5.9 \\
BMB~(ours)                           &  \textbf{35.2}\small{~$\uparrow$4.3}  &  \textbf{51.2}\small{~$\uparrow$0.9}  &  \textbf{29.0}\small{~$\uparrow$6.8}  &  \textbf{12.0}\small{~$\uparrow$6.1} \\ \bottomrule 
\end{tabular}
} \\
\vspace{0.1cm}
{(b) ImageNet-LT 50$\%$ labeled subset} \\
\vspace{0.2cm}
\caption{Results on ImageNet-LT (a) 20$\%$ labeled dataset and (b) 50$\%$ labeled dataset.  For the 20$\%$ subset, we classify classes with more than 20 training samples as many-shot, fewer than 4 samples as few-shot, the remaining as medium-shot. The partition intervals for the 50$\%$ subset can be found in the header of subtable (b).}
\label{tab:exp_imagenet-lt}
\end{table}
\fakeparagraph{Network architecture.} 
On the ImageNet127 and ImageNet-LT datasets, we use ResNet50~\cite{resnet} as the encoder, and train it from scratch. When conducting experiments on CIFAR10-LT and CIFAR100-LT, we follow the common practice in previous works~\cite{DARP,wei2021crest,fan2022cossl}, and use the randomly initialized WideResNet-28~\cite{wider-resnet} as the encoder. In all cases, the base classifier and the auxiliary classifier are both single-layer linear classifiers.

\fakeparagraph{Training setups.}
For a fair comparison, we keep the training and evaluation setups identical to those in previous works~\cite{DARP,wei2021crest,fan2022cossl}. Specifically, 
we train the models for 500 epochs on ImageNet127, CIFAR10-LT and CIFAR100-LT, and 300 epochs on ImageNet-LT, with each epoch consisting of 500 iterations. For all datasets, we utilize Adam~\cite{Kingma2014AdamAM} optimizer with a constant learning rate of 0.002 without any scheduling. The batch size is 64 for both labeled and unlabeled data across all datasets. The size of the class-rebalanced memory bank is 128 for CIFAR10-LT, 256 for CIFAR100-LT and ImageNet127, and 1024 for ImageNet-LT. At each training step, we select a certain proportion of features from the memory and use their pseudo-labels for training. Specifically, we select 50$\%$ of features on CIFAR and ImageNet127, and 25$\%$ on ImageNet-LT. More detailed hyperparameters setting can be found in the supplementary~\ref{sec:sup_implementation_details}.

\fakeparagraph{Evaluation metrics.}
For ImageNet127, we report the averaged class recall of the last 20 epochs due to the imbalanced test set. For ImageNet-LT, we save the checkpoint that achieves the best accuracy on the on validation set, and report its accuracy on a hold-out test set. For CIFAR, we report the averaged test accuracy of the last 20 epochs, following the approaches in~\cite{wider-resnet, fan2022cossl}. It is worth noting that we evaluate the performance using an exponential moving average of the parameters over training with a decay rate of 0.999, as is common practice in~\cite{berthelot2019mixmatch,fan2022cossl,DARP}.

\subsection{Main Results}
\fakeparagraph{ImageNet-LT.}
We conduct experiments on the 20$\%$ and 50$\%$ labeled subsets of the original dataset. In the 20$\%$ subset, there has only one labeled sample in most scarce class, leading to an extremely difficult task. 
To gain a deeper understanding of each method, following previous works~\cite{imagenet-lt, li2022targeted}, we not only consider the overall top-1 accuracy across all classes but also evaluate the accuracy of three disjoint subsets: many-shot, medium-shot, and few-shot classes. The experimental results and partitioning rules for these subsets can be found in~\cref{tab:exp_imagenet-lt}. 
We can observe that \system achieves an overall accuracy that exceeds other methods by 2.8$\%$ and 4.3$\%$ on the 20$\%$ and 50$\%$ subsets, respectively. 

\fakeparagraph{ImageNet127.}
To remain consistent with prior work~\cite{fan2022cossl} and save computational resources, we adopt the approach described in~\cite{Chrabaszcz2017ADV} to downsample the images in ImageNet
 to resolutions of $32\times32$ or $64\times64$, which was also employed by~\cite{fan2022cossl}. This yield a downsampled variant of ImageNet127 that we used for our experiments. The outcomes obtained under various resolutions and labeled subsets are summarized in~\cref{tab:exp_imagenet127}. We can observe that our method outperforms other methods significantly in all settings, particularly in the 1$\%$ subset with a $64\times64$ resolution, where we surpass the second-best method by 8.2$\%$.

\begin{table}[]
\centering
\resizebox{0.8\columnwidth}{!}{%
\renewcommand\arraystretch{1.2}
\begin{tabular}{lccccc}
\toprule
& \multicolumn{4}{c}{ImageNet127}    \\ \cmidrule(lr){2-5} 
& \multicolumn{2}{c}{1\%}  & \multicolumn{2}{c}{10\%}  \\ \cmidrule(lr){2-3} \cmidrule(lr){4-5}
& \(32\times32\) &  \(64\times64\)  & \(32\times32\) & \(64\times64\)  \\ \cmidrule(lr){2-3} \cmidrule(lr){4-5}
Vanilla~\cite{resnet}            &  8.4  &  11.1 &  29.4  &  38.5  \\
FixMatch~\cite{sohn2020fixmatch} &  9.9  &  15.2 &  29.8  &  43.6  \\
CReST~\cite{wei2021crest}        &  8.5  &  10.3 &  28.1  &  38.8  \\
DARP~\cite{DARP}                 &  10.0 &  16.6 &  30.9  &  43.2  \\
CoSSL~\cite{fan2022cossl}        &  14.9 &  19.3 &  44.0  &  52.5  \\ \cmidrule(lr){1-5}
BMB~(ours)                       &  \textbf{18.4} &  \textbf{27.5} &  \textbf{46.8}  &  \textbf{56.4}  \\ \bottomrule
\end{tabular}%
}
\smallskip
\caption{Averaged class recall ($\%$) under different input resolutions and different scales of labeled data. We reproduce all the other algorithms using the same codebase released by~\cite{fan2022cossl} for a fair comparison. The best results are in bold.}
\label{tab:exp_imagenet127}
\end{table}

\fakeparagraph{CIFAR10-LT and CIFAR100-LT.}
We also conduct experiments on CIFAR~\cite{cifar}, assuming that the labeled and unlabeled datasets share the same distribution, \ie~$\gamma=\gamma_l=\gamma_u$. We report results with $\gamma=20$ for CIFAR10-LT and $\gamma=10$ for CIFAR100-LT. We run each experiment with three random seeds and report the means and standard deviations in~\cref{tab:exp_cifar_main}. Our \system show the best accuracy comparing with previous state-of-the-art methods, and achieve an improvement of 1.0$\%$ on CIFAR100-LT with $\gamma=10$.
\begin{table}[]
\centering
\resizebox{0.85\columnwidth}{!}{%
\renewcommand\arraystretch{1.2}
\begin{tabular}{lcc}
\toprule
& CIFAR10 ($\gamma=20$) & CIFAR100 ($\gamma=10$) \\ \cmidrule(lr){2-3}
Vanilla~\cite{wider-resnet}        &  $76.2_{\pm{0.51}}$  &   $42.3_{\pm{0.95}}$   \\
FixMatch~\cite{sohn2020fixmatch}   &  $87.7_{\pm{0.34}}$  &   $55.3_{\pm{0.20}}$   \\
CReST+~\cite{wei2021crest}         &  $86.9_{\pm{0.36}}$  &   $54.8_{\pm{0.12}}$   \\
DARP~\cite{DARP}                   &  $88.1_{\pm{0.23}}$  &   $54.5_{\pm{0.18}}$   \\
CoSSL~\cite{fan2022cossl}          &  $89.8_{\pm{0.30}}$  &   $58.4_{\pm{0.16}}$   \\
\cmidrule(lr){1-3}
BMB~(ours)                          &  $\textbf{90.2}_{\pm{0.40}}$  &  $\textbf{59.4}_{\pm{0.01}}$   \\ \bottomrule
\end{tabular}%
}
\smallskip
\caption{Top-1 accuracy (\%) on CIFAR10-LT and CIFAR100-LT with different imbalance ratio, and the test dataset is remain balanced. We reproduce all the algorithms using the codebase released by~\cite{fan2022cossl} for a fair comparison.}
\label{tab:exp_cifar_main}
\end{table}

\fakeparagraph{Results under mismatched data distributions.}
In more realistic scenarios, labeled and unlabeled data may not share the same distribution, making it crucial to test method effectiveness when $\gamma_l\neq\gamma_u$.
The ImageNet-LT and ImageNet127 datasets are unsuitable for such testing since their imbalance ratios are fixed. Therefore, we sample a subset from the original ImageNet dataset~\cite{Russakovsky2014ImageNetLS} using the same method as for constructing the long-tailed CIFAR. Specifically, we set $N_1=600$, $M_1=300$ and fix $\gamma_l=50$ while $\gamma_u$ varies between 1, 20 and 100.
The experimental results presented in ~\cref{tab:exp_distribution_mismatch} demonstrate that our method achieves the highest accuracy across different settings. We attribute this to our method's ability to make no assumptions about the distribution of unlabeled data and estimate it through an effective method.

\begin{table}[]
\centering
\resizebox{0.65\columnwidth}{!}{%
\renewcommand\arraystretch{1.2}
\begin{tabular}{lccc}
\toprule
   & \multicolumn{3}{c}{ImageNet ($\gamma_l=50)$}  \\ \cmidrule(lr){2-4}
   & $\gamma_u=1$ & $\gamma_u=20$ & $\gamma_u=100$ \\ \cmidrule(lr){2-4}
Vanilla~\cite{wider-resnet}       &  33.2 &  34.7 &   34.1 \\
FixMatch~\cite{sohn2020fixmatch}  &  38.9 &  39.5 &   37.8 \\
CoSSL~\cite{fan2022cossl}         &  39.6 &  39.3 &   38.1 \\
CReST+~\cite{wei2021crest}        &  39.5 &  39.8 &   40.3 \\
DARP~\cite{DARP}                  &  46.7 &  46.7 &   46.9 \\ \cmidrule(lr){1-4}
BMB~(ours)                        &  49.9 &  48.7 &   48.4 \\ \bottomrule
\end{tabular}%
}
\smallskip
\caption{Top-1 ($\%$) accuracy on the long-tailed version of ImageNet dataset~\cite{Russakovsky2014ImageNetLS}, where distributions of the labeled and unlabeled datasets are mismatched.}
\label{tab:exp_distribution_mismatch}
\end{table}

\subsection{Ablation Studies}

\begin{figure*}[h]
  \centering
  \includegraphics[width=0.85\linewidth]{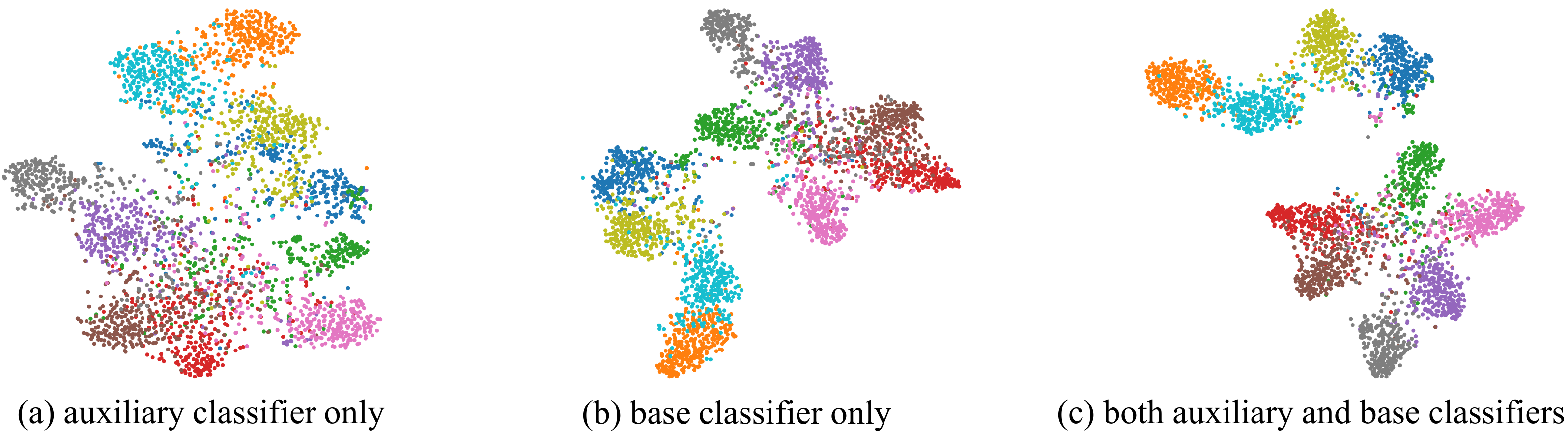}
  \caption{T-SNE~\cite{t-sne} visualization of the extracted representations learned under different classifier configurations: (a) only the auxiliary classifier, (b) only the base classifier and (c) both the auxiliary and the base classifier are included in the training process of \system~(the default configuration of \system).}
  \label{fig:t-sne}
\end{figure*}

To investigate the importance of different components and the settings of key hyperparameters, we conduct ablation experiments and related discussions in this section. The implementation details can be found in supplementary~\ref{sec:sup_implementation_details}.

\fakeparagraph{Main components of \system.} 
There are two main components in \system: the class-rebalanced feature memory bank and the adaptive weighting module. To investigate the effectiveness of each component, we gradually add each one and present the experimental results in~\cref{tab:exp_ablate_componet}. We observe that our method only achieves a modest improvement of 1.0$\%$ over the baseline when using the memory bank alone. However, when the adaptive weighting module is attached to the memory, the accuracy is further improved by 2.2$\%$. 
\begin{table}[]
\centering
\resizebox{0.48\columnwidth}{!}{%
\renewcommand\arraystretch{1.2}
\begin{tabular}{ccc}
\toprule
adaW & memory & accuracy (\%) \\ \cmidrule(lr){1-2} \cmidrule(lr){3-3}
             &                &    56.2   \\
             &  \Checkmark    &    57.2   \\
\Checkmark   &  \Checkmark    &    59.4   \\  \bottomrule
\end{tabular}%
}
\smallskip
\caption{We incrementally introduced each module of \system to evaluate their individual importance.}
\label{tab:exp_ablate_componet}
\end{table}

\begin{figure}
  \centering
  \includegraphics[width=0.85\linewidth]{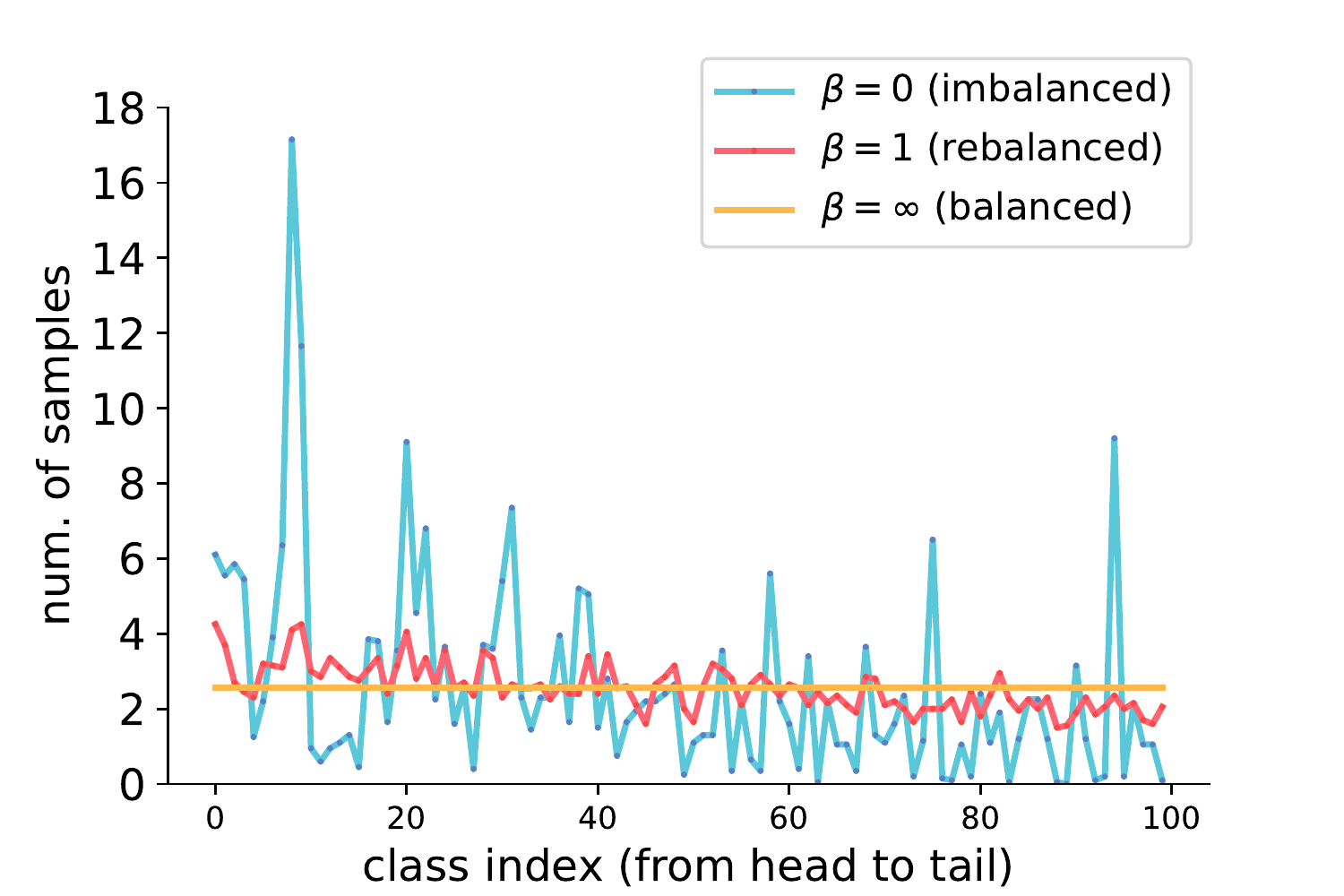}
  \caption{The distribution of samples from different classes in the memory bank. The larger the value of $\beta$, the more balanced the distribution will be.}
  \label{fig:in_memory_distribution}
\end{figure}

\fakeparagraph{Rebalancing degree of the memory bank.} 
In the maintenance and updating of the memory bank, there is a critical parameter that controls the degree of rebalancing, namely the coefficient $\beta$ in~\cref{eq:put_prob} and ~\cref{eq:remove_prob}. As we can see from the equations, a larger value of $\beta$ can lead to a more balanced distribution of features from different classes in the memory bank. When $\beta$ equals zero, the maintenance of the memory bank becomes random, and all features are added to or removed from the memory bank with equal probability, regardless of their category. We visualize the distribution of data in the memory bank under different values of $\beta$ in~\cref{fig:in_memory_distribution}. When $\beta=0$, the data in the memory bank exhibits a imbalanced distribution, this is because the unlabeled data is inherently imbalanced. When $\beta=1$, the imbalance is significantly alleviated and the distribution is very close to the ideal balanced distribution~(when $\beta=\infty$). This indicates the our algorithm is effective in rebalancing the imbalanced data. 

\begin{figure}
  \centering
  \includegraphics[width=0.7\linewidth]{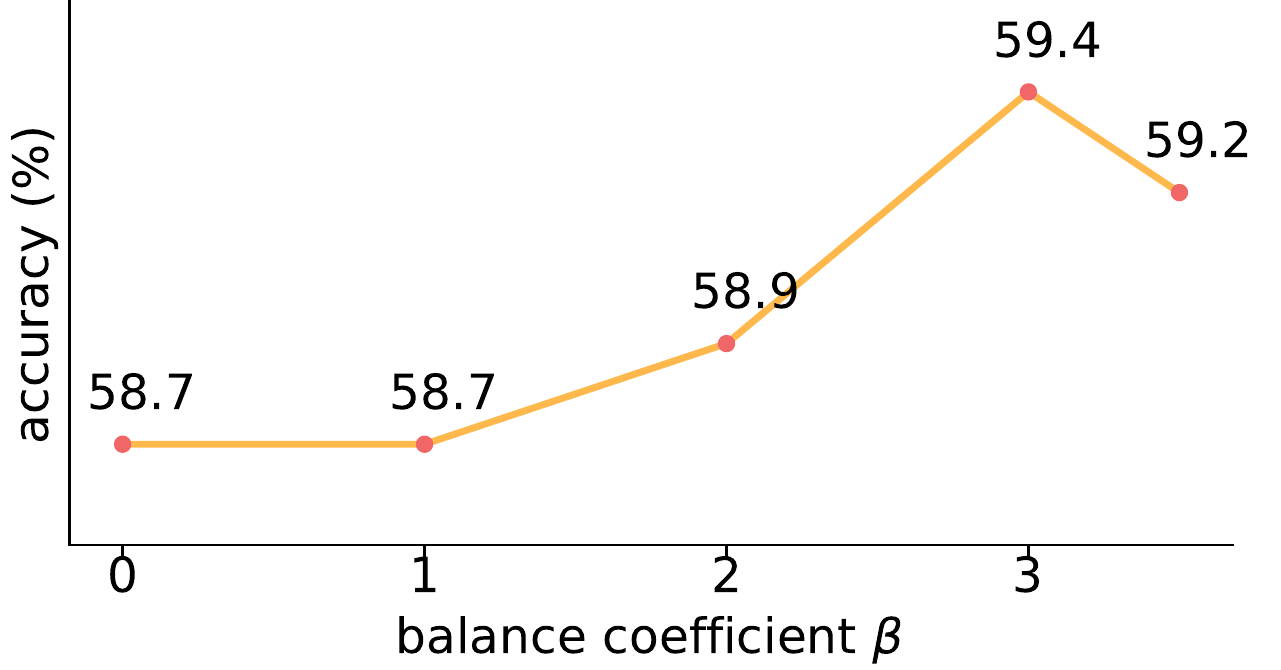}
  \caption{The test accuracy under different $\beta$ values, indicating how the balance degree effects the model's performance.}
  \label{fig:bp_acc}
\end{figure}

To further investigate how the in-memory data distribution affects model performance, we present the accuracy of the model at various values of $\beta$ in ~\cref{fig:bp_acc}. It can be observed that the model performs poorly when the memory bank is imbalanced, and the accuracy increases as $\beta$ increases. However, when $\beta$ becomes too large, the performance starts to decline. We speculate that this is due to an excessive emphasis on data balance, which may affect the updating rate of data. The results indicate that maintaining a moderately balanced memory bank is necessary.

\fakeparagraph{Necessity of the base classifier.}
\label{sec:ablate_classfiers}
Two separate classifiers are used in the training process of \system: a class-rebalanced auxiliary classifier and a vanilla base classifier. During inference, only the auxiliary classifier is utilized while the base classifier is discarded. To validate the need for the base classifier, we employ t-SNE~\cite{t-sne} to visualize the representations extracted by the encoder trained with different classifier configurations in \cref{fig:t-sne}. As depicted in~\cref{fig:t-sne}(a), the quality of the features extracted by the encoder is poor when only the auxiliary classifier is utilized. However, when the base classifier is incorporated on top of it (\cref{fig:t-sne}(c)), the extracted features are significantly enhanced. Meanwhile, as shown in~\cref{fig:t-sne}(b), the quality of the extracted features is also decent when only the base classifier is used, which is in line with the findings in~\cite{kang2019decoupling,fan2022cossl} that the imbalanced data has little effect on the encoder.

\section{Conclusion}
This work delved into the challenging and under-explored problem of class-imbalanced SSL. We proposed a novel approach named \system, which centers on an online-updated memory bank. The memory caches the historical features and their corresponding pseudo labels, and a crafted algorithm is designed to ensure the inside data distribution to be class-rebalanced. Building on this well-curated memory, we apply a re-sampling strategy at the feature level to mitigate the impact of imbalanced training data. To better re-calibrate the classifier and ensure the memory bank being well-initialized and maintained, we also introduce an adaptive weighting module to assist the memory bank. With all the crafted components working in synergy, \system successfully rebalanced the learning process of the classifier, leading to state-of-the-art performance across multiple imbalanced SSL benchmarks.

\bibliographystyle{ACM-Reference-Format}
\bibliography{ref}


\begin{thebibliography}{36}

%%% the Web address.

\ifx \showCODEN    \undefined \def \showCODEN     #1{\unskip}     \fi
\ifx \showDOI      \undefined \def \showDOI       #1{#1}\fi
\ifx \showISBNx    \undefined \def \showISBNx     #1{\unskip}     \fi
\ifx \showISBNxiii \undefined \def \showISBNxiii  #1{\unskip}     \fi
\ifx \showISSN     \undefined \def \showISSN      #1{\unskip}     \fi
\ifx \showLCCN     \undefined \def \showLCCN      #1{\unskip}     \fi
\ifx \shownote     \undefined \def \shownote      #1{#1}          \fi
\ifx \showarticletitle \undefined \def \showarticletitle #1{#1}   \fi
\ifx \showURL      \undefined \def \showURL       {\relax}        \fi
% invisible to TeX
\providecommand\bibfield[2]{#2}
\providecommand\bibinfo[2]{#2}
\providecommand\natexlab[1]{#1}
\providecommand\showeprint[2][]{arXiv:#2}

\bibitem[Berthelot et~al\mbox{.}(2019)]%
        {berthelot2019mixmatch}
\bibfield{author}{\bibinfo{person}{David Berthelot}, \bibinfo{person}{Nicholas
  Carlini}, \bibinfo{person}{Ian Goodfellow}, \bibinfo{person}{Nicolas
  Papernot}, \bibinfo{person}{Avital Oliver}, {and} \bibinfo{person}{Colin~A
  Raffel}.} \bibinfo{year}{2019}\natexlab{}.
\newblock \showarticletitle{Mixmatch: A holistic approach to semi-supervised
  learning}. In \bibinfo{booktitle}{\emph{NeurIPS}}.
\newblock


\bibitem[Buda et~al\mbox{.}(2018)]%
        {buda2018systematic}
\bibfield{author}{\bibinfo{person}{Mateusz Buda}, \bibinfo{person}{Atsuto
  Maki}, {and} \bibinfo{person}{Maciej~A Mazurowski}.}
  \bibinfo{year}{2018}\natexlab{}.
\newblock \showarticletitle{A systematic study of the class imbalance problem
  in convolutional neural networks}.
\newblock \bibinfo{journal}{\emph{Neural networks}}  \bibinfo{volume}{106}
  (\bibinfo{year}{2018}).
\newblock


\bibitem[Cao et~al\mbox{.}(2019)]%
        {cao2019learning}
\bibfield{author}{\bibinfo{person}{Kaidi Cao}, \bibinfo{person}{Colin Wei},
  \bibinfo{person}{Adrien Gaidon}, \bibinfo{person}{Nikos Arechiga}, {and}
  \bibinfo{person}{Tengyu Ma}.} \bibinfo{year}{2019}\natexlab{}.
\newblock \showarticletitle{Learning imbalanced datasets with
  label-distribution-aware margin loss}. In
  \bibinfo{booktitle}{\emph{NeurIPS}}.
\newblock


\bibitem[Chawla et~al\mbox{.}(2002)]%
        {chawla2002smote}
\bibfield{author}{\bibinfo{person}{Nitesh~V Chawla}, \bibinfo{person}{Kevin~W
  Bowyer}, \bibinfo{person}{Lawrence~O Hall}, {and} \bibinfo{person}{W~Philip
  Kegelmeyer}.} \bibinfo{year}{2002}\natexlab{}.
\newblock \showarticletitle{SMOTE: synthetic minority over-sampling technique}.
\newblock \bibinfo{journal}{\emph{JAIR}}  \bibinfo{volume}{16}
  (\bibinfo{year}{2002}).
\newblock


\bibitem[Chrabaszcz et~al\mbox{.}(2017)]%
        {Chrabaszcz2017ADV}
\bibfield{author}{\bibinfo{person}{Patryk Chrabaszcz}, \bibinfo{person}{Ilya
  Loshchilov}, {and} \bibinfo{person}{Frank Hutter}.}
  \bibinfo{year}{2017}\natexlab{}.
\newblock \showarticletitle{A Downsampled Variant of ImageNet as an Alternative
  to the CIFAR datasets}.
\newblock \bibinfo{journal}{\emph{ArXiv}}  \bibinfo{volume}{abs/1707.08819}
  (\bibinfo{year}{2017}).
\newblock


\bibitem[Cui et~al\mbox{.}(2019)]%
        {cui2019class}
\bibfield{author}{\bibinfo{person}{Yin Cui}, \bibinfo{person}{Menglin Jia},
  \bibinfo{person}{Tsung-Yi Lin}, \bibinfo{person}{Yang Song}, {and}
  \bibinfo{person}{Serge Belongie}.} \bibinfo{year}{2019}\natexlab{}.
\newblock \showarticletitle{Class-balanced loss based on effective number of
  samples}. In \bibinfo{booktitle}{\emph{CVPR}}.
\newblock


\bibitem[Fan et~al\mbox{.}(2022)]%
        {fan2022cossl}
\bibfield{author}{\bibinfo{person}{Yue Fan}, \bibinfo{person}{Dengxin Dai},
  \bibinfo{person}{Anna Kukleva}, {and} \bibinfo{person}{Bernt Schiele}.}
  \bibinfo{year}{2022}\natexlab{}.
\newblock \showarticletitle{Cossl: Co-learning of representation and classifier
  for imbalanced semi-supervised learning}. In
  \bibinfo{booktitle}{\emph{CVPR}}.
\newblock


\bibitem[He and Garcia(2009)]%
        {he2009learning}
\bibfield{author}{\bibinfo{person}{Haibo He} {and} \bibinfo{person}{Edwardo~A
  Garcia}.} \bibinfo{year}{2009}\natexlab{}.
\newblock \showarticletitle{Learning from imbalanced data}.
\newblock \bibinfo{journal}{\emph{TKDE}} \bibinfo{volume}{21},
  \bibinfo{number}{9} (\bibinfo{year}{2009}).
\newblock


\bibitem[He et~al\mbox{.}(2021)]%
        {He2021RethinkingRI}
\bibfield{author}{\bibinfo{person}{Ju He}, \bibinfo{person}{Adam Kortylewski},
  \bibinfo{person}{Shaokang Yang}, \bibinfo{person}{Shuai Liu},
  \bibinfo{person}{Cheng Yang}, \bibinfo{person}{Changhu Wang}, {and}
  \bibinfo{person}{Alan~Loddon Yuille}.} \bibinfo{year}{2021}\natexlab{}.
\newblock \showarticletitle{Rethinking Re-Sampling in Imbalanced
  Semi-Supervised Learning}.
\newblock \bibinfo{journal}{\emph{ArXiv}}  \bibinfo{volume}{abs/2106.00209}
  (\bibinfo{year}{2021}).
\newblock


\bibitem[He et~al\mbox{.}(2016)]%
        {resnet}
\bibfield{author}{\bibinfo{person}{Kaiming He}, \bibinfo{person}{Xiangyu
  Zhang}, \bibinfo{person}{Shaoqing Ren}, {and} \bibinfo{person}{Jian Sun}.}
  \bibinfo{year}{2016}\natexlab{}.
\newblock \showarticletitle{Deep residual learning for image recognition}. In
  \bibinfo{booktitle}{\emph{CVPR}}.
\newblock


\bibitem[Huh et~al\mbox{.}(2016)]%
        {imagenet127}
\bibfield{author}{\bibinfo{person}{Minyoung Huh}, \bibinfo{person}{Pulkit
  Agrawal}, {and} \bibinfo{person}{Alexei~A Efros}.}
  \bibinfo{year}{2016}\natexlab{}.
\newblock \showarticletitle{What makes ImageNet good for transfer learning?}
\newblock \bibinfo{journal}{\emph{arXiv preprint arXiv:1608.08614}}
  (\bibinfo{year}{2016}).
\newblock


\bibitem[Kang et~al\mbox{.}(2020)]%
        {kang2019decoupling}
\bibfield{author}{\bibinfo{person}{Bingyi Kang}, \bibinfo{person}{Saining Xie},
  \bibinfo{person}{Marcus Rohrbach}, \bibinfo{person}{Zhicheng Yan},
  \bibinfo{person}{Albert Gordo}, \bibinfo{person}{Jiashi Feng}, {and}
  \bibinfo{person}{Yannis Kalantidis}.} \bibinfo{year}{2020}\natexlab{}.
\newblock \showarticletitle{Decoupling representation and classifier for
  long-tailed recognition}. In \bibinfo{booktitle}{\emph{ICLR}}.
\newblock


\bibitem[Kim et~al\mbox{.}(2020)]%
        {DARP}
\bibfield{author}{\bibinfo{person}{Jaehyung Kim}, \bibinfo{person}{Youngbum
  Hur}, \bibinfo{person}{Sejun Park}, \bibinfo{person}{Eunho Yang},
  \bibinfo{person}{Sung~Ju Hwang}, {and} \bibinfo{person}{Jinwoo Shin}.}
  \bibinfo{year}{2020}\natexlab{}.
\newblock \showarticletitle{Distribution aligning refinery of pseudo-label for
  imbalanced semi-supervised learning}. In \bibinfo{booktitle}{\emph{NeurIPS}}.
\newblock


\bibitem[Kingma and Ba(2014)]%
        {Kingma2014AdamAM}
\bibfield{author}{\bibinfo{person}{Diederik~P. Kingma} {and}
  \bibinfo{person}{Jimmy Ba}.} \bibinfo{year}{2014}\natexlab{}.
\newblock \showarticletitle{Adam: A Method for Stochastic Optimization}.
\newblock \bibinfo{journal}{\emph{CoRR}} (\bibinfo{year}{2014}).
\newblock


\bibitem[Krizhevsky et~al\mbox{.}(2009)]%
        {cifar}
\bibfield{author}{\bibinfo{person}{Alex Krizhevsky}, \bibinfo{person}{Geoffrey
  Hinton}, {et~al\mbox{.}}} \bibinfo{year}{2009}\natexlab{}.
\newblock \showarticletitle{Learning multiple layers of features from tiny
  images}.
\newblock  (\bibinfo{year}{2009}).
\newblock


\bibitem[Laine and Aila(2017)]%
        {temporal-ensembling}
\bibfield{author}{\bibinfo{person}{Samuli Laine} {and} \bibinfo{person}{Timo
  Aila}.} \bibinfo{year}{2017}\natexlab{}.
\newblock \showarticletitle{Temporal ensembling for semi-supervised learning}.
  In \bibinfo{booktitle}{\emph{ICLR}}.
\newblock


\bibitem[Lee et~al\mbox{.}(2013)]%
        {pseudo-label}
\bibfield{author}{\bibinfo{person}{Dong-Hyun Lee} {et~al\mbox{.}}}
  \bibinfo{year}{2013}\natexlab{}.
\newblock \showarticletitle{Pseudo-label: The simple and efficient
  semi-supervised learning method for deep neural networks}. In
  \bibinfo{booktitle}{\emph{ICML workshop}}.
\newblock


\bibitem[Lee et~al\mbox{.}(2021)]%
        {lee2021abc}
\bibfield{author}{\bibinfo{person}{Hyuck Lee}, \bibinfo{person}{Seungjae Shin},
  {and} \bibinfo{person}{Heeyoung Kim}.} \bibinfo{year}{2021}\natexlab{}.
\newblock \showarticletitle{ABC: Auxiliary Balanced Classifier for
  Class-imbalanced Semi-supervised Learning}. In
  \bibinfo{booktitle}{\emph{NeurIPS}}.
\newblock


\bibitem[Li et~al\mbox{.}(2022)]%
        {li2022targeted}
\bibfield{author}{\bibinfo{person}{Tianhong Li}, \bibinfo{person}{Peng Cao},
  \bibinfo{person}{Yuan Yuan}, \bibinfo{person}{Lijie Fan},
  \bibinfo{person}{Yuzhe Yang}, \bibinfo{person}{Rogerio~S Feris},
  \bibinfo{person}{Piotr Indyk}, {and} \bibinfo{person}{Dina Katabi}.}
  \bibinfo{year}{2022}\natexlab{}.
\newblock \showarticletitle{Targeted supervised contrastive learning for
  long-tailed recognition}. In \bibinfo{booktitle}{\emph{CVPR}}.
\newblock


\bibitem[Lin et~al\mbox{.}(2017)]%
        {lin2017focal}
\bibfield{author}{\bibinfo{person}{Tsung-Yi Lin}, \bibinfo{person}{Priya
  Goyal}, \bibinfo{person}{Ross Girshick}, \bibinfo{person}{Kaiming He}, {and}
  \bibinfo{person}{Piotr Doll{\'a}r}.} \bibinfo{year}{2017}\natexlab{}.
\newblock \showarticletitle{Focal loss for dense object detection}. In
  \bibinfo{booktitle}{\emph{ICCV}}.
\newblock


\bibitem[Liu et~al\mbox{.}(2019)]%
        {imagenet-lt}
\bibfield{author}{\bibinfo{person}{Ziwei Liu}, \bibinfo{person}{Zhongqi Miao},
  \bibinfo{person}{Xiaohang Zhan}, \bibinfo{person}{Jiayun Wang},
  \bibinfo{person}{Boqing Gong}, {and} \bibinfo{person}{Stella~X Yu}.}
  \bibinfo{year}{2019}\natexlab{}.
\newblock \showarticletitle{Large-scale long-tailed recognition in an open
  world}. In \bibinfo{booktitle}{\emph{CVPR}}.
\newblock


\bibitem[Menon et~al\mbox{.}(2021)]%
        {LA}
\bibfield{author}{\bibinfo{person}{Aditya~Krishna Menon},
  \bibinfo{person}{Sadeep Jayasumana}, \bibinfo{person}{Ankit~Singh Rawat},
  \bibinfo{person}{Himanshu Jain}, \bibinfo{person}{Andreas Veit}, {and}
  \bibinfo{person}{Sanjiv Kumar}.} \bibinfo{year}{2021}\natexlab{}.
\newblock \showarticletitle{Long-tail learning via logit adjustment}. In
  \bibinfo{booktitle}{\emph{ICLR}}.
\newblock


\bibitem[Miyato et~al\mbox{.}(2018)]%
        {vat}
\bibfield{author}{\bibinfo{person}{Takeru Miyato}, \bibinfo{person}{Shin-ichi
  Maeda}, \bibinfo{person}{Masanori Koyama}, {and} \bibinfo{person}{Shin
  Ishii}.} \bibinfo{year}{2018}\natexlab{}.
\newblock \showarticletitle{Virtual adversarial training: a regularization
  method for supervised and semi-supervised learning}.
\newblock \bibinfo{journal}{\emph{TPAMI}} \bibinfo{volume}{41},
  \bibinfo{number}{8} (\bibinfo{year}{2018}).
\newblock


\bibitem[Oh et~al\mbox{.}(2022)]%
        {DASO}
\bibfield{author}{\bibinfo{person}{Youngtaek Oh}, \bibinfo{person}{Dong-Jin
  Kim}, {and} \bibinfo{person}{In~So Kweon}.} \bibinfo{year}{2022}\natexlab{}.
\newblock \showarticletitle{DASO: Distribution-aware semantics-oriented
  pseudo-label for imbalanced semi-supervised learning}. In
  \bibinfo{booktitle}{\emph{CVPR}}.
\newblock


\bibitem[Oliver et~al\mbox{.}(2018)]%
        {wider-resnet}
\bibfield{author}{\bibinfo{person}{Avital Oliver}, \bibinfo{person}{Augustus
  Odena}, \bibinfo{person}{Colin Raffel}, \bibinfo{person}{Ekin~Dogus Cubuk},
  {and} \bibinfo{person}{Ian~J. Goodfellow}.} \bibinfo{year}{2018}\natexlab{}.
\newblock \showarticletitle{Realistic Evaluation of Deep Semi-Supervised
  Learning Algorithms}. In \bibinfo{booktitle}{\emph{NeurIPS}}.
\newblock


\bibitem[Russakovsky et~al\mbox{.}(2014)]%
        {Russakovsky2014ImageNetLS}
\bibfield{author}{\bibinfo{person}{Olga Russakovsky}, \bibinfo{person}{Jia
  Deng}, \bibinfo{person}{Hao Su}, \bibinfo{person}{Jonathan Krause},
  \bibinfo{person}{Sanjeev Satheesh}, \bibinfo{person}{Sean Ma},
  \bibinfo{person}{Zhiheng Huang}, \bibinfo{person}{Andrej Karpathy},
  \bibinfo{person}{Aditya Khosla}, \bibinfo{person}{Michael~S. Bernstein},
  \bibinfo{person}{Alexander~C. Berg}, {and} \bibinfo{person}{Li Fei-Fei}.}
  \bibinfo{year}{2014}\natexlab{}.
\newblock \showarticletitle{ImageNet Large Scale Visual Recognition Challenge}.
\newblock \bibinfo{journal}{\emph{IJCV}}  \bibinfo{volume}{115}
  (\bibinfo{year}{2014}).
\newblock


\bibitem[Sohn et~al\mbox{.}(2020)]%
        {sohn2020fixmatch}
\bibfield{author}{\bibinfo{person}{Kihyuk Sohn}, \bibinfo{person}{David
  Berthelot}, \bibinfo{person}{Nicholas Carlini}, \bibinfo{person}{Zizhao
  Zhang}, \bibinfo{person}{Han Zhang}, \bibinfo{person}{Colin~A Raffel},
  \bibinfo{person}{Ekin~Dogus Cubuk}, \bibinfo{person}{Alexey Kurakin}, {and}
  \bibinfo{person}{Chun-Liang Li}.} \bibinfo{year}{2020}\natexlab{}.
\newblock \showarticletitle{Fixmatch: Simplifying semi-supervised learning with
  consistency and confidence}. In \bibinfo{booktitle}{\emph{NeurIPS}}.
\newblock


\bibitem[Tan et~al\mbox{.}(2020)]%
        {Tan2020EqualizationLF}
\bibfield{author}{\bibinfo{person}{Jingru Tan}, \bibinfo{person}{Changbao
  Wang}, \bibinfo{person}{Buyu Li}, \bibinfo{person}{Quanquan Li},
  \bibinfo{person}{Wanli Ouyang}, \bibinfo{person}{Changqing Yin}, {and}
  \bibinfo{person}{Junjie Yan}.} \bibinfo{year}{2020}\natexlab{}.
\newblock \showarticletitle{Equalization Loss for Long-Tailed Object
  Recognition}. In \bibinfo{booktitle}{\emph{CVPR}}.
\newblock


\bibitem[Tarvainen and Valpola(2017)]%
        {mean-teahcer}
\bibfield{author}{\bibinfo{person}{Antti Tarvainen} {and}
  \bibinfo{person}{Harri Valpola}.} \bibinfo{year}{2017}\natexlab{}.
\newblock \showarticletitle{Mean teachers are better role models:
  Weight-averaged consistency targets improve semi-supervised deep learning
  results}. In \bibinfo{booktitle}{\emph{NeurIPS}}.
\newblock


\bibitem[van~der Maaten and Hinton(2008)]%
        {t-sne}
\bibfield{author}{\bibinfo{person}{Laurens van~der Maaten} {and}
  \bibinfo{person}{Geoffrey~E. Hinton}.} \bibinfo{year}{2008}\natexlab{}.
\newblock \showarticletitle{Visualizing Data using t-SNE}.
\newblock \bibinfo{journal}{\emph{JMLR}} (\bibinfo{year}{2008}).
\newblock


\bibitem[Wang et~al\mbox{.}(2021)]%
        {wang2021contrastive}
\bibfield{author}{\bibinfo{person}{Peng Wang}, \bibinfo{person}{Kai Han},
  \bibinfo{person}{Xiu-Shen Wei}, \bibinfo{person}{Lei Zhang}, {and}
  \bibinfo{person}{Lei Wang}.} \bibinfo{year}{2021}\natexlab{}.
\newblock \showarticletitle{Contrastive learning based hybrid networks for
  long-tailed image classification}. In \bibinfo{booktitle}{\emph{CVPR}}.
\newblock


\bibitem[Wei et~al\mbox{.}(2021)]%
        {wei2021crest}
\bibfield{author}{\bibinfo{person}{Chen Wei}, \bibinfo{person}{Kihyuk Sohn},
  \bibinfo{person}{Clayton Mellina}, \bibinfo{person}{Alan Yuille}, {and}
  \bibinfo{person}{Fan Yang}.} \bibinfo{year}{2021}\natexlab{}.
\newblock \showarticletitle{Crest: A class-rebalancing self-training framework
  for imbalanced semi-supervised learning}. In
  \bibinfo{booktitle}{\emph{CVPR}}.
\newblock


\bibitem[Xie et~al\mbox{.}(2020a)]%
        {uda}
\bibfield{author}{\bibinfo{person}{Qizhe Xie}, \bibinfo{person}{Zihang Dai},
  \bibinfo{person}{Eduard Hovy}, \bibinfo{person}{Thang Luong}, {and}
  \bibinfo{person}{Quoc Le}.} \bibinfo{year}{2020}\natexlab{a}.
\newblock \showarticletitle{Unsupervised data augmentation for consistency
  training}. In \bibinfo{booktitle}{\emph{NeurIPS}}.
\newblock


\bibitem[Xie et~al\mbox{.}(2020b)]%
        {noisystudent}
\bibfield{author}{\bibinfo{person}{Qizhe Xie}, \bibinfo{person}{Minh-Thang
  Luong}, \bibinfo{person}{Eduard Hovy}, {and} \bibinfo{person}{Quoc~V Le}.}
  \bibinfo{year}{2020}\natexlab{b}.
\newblock \showarticletitle{Self-training with noisy student improves imagenet
  classification}. In \bibinfo{booktitle}{\emph{CVPR}}.
\newblock


\bibitem[Zhou et~al\mbox{.}(2020)]%
        {zhou2020bbn}
\bibfield{author}{\bibinfo{person}{Boyan Zhou}, \bibinfo{person}{Quan Cui},
  \bibinfo{person}{Xiu-Shen Wei}, {and} \bibinfo{person}{Zhao-Min Chen}.}
  \bibinfo{year}{2020}\natexlab{}.
\newblock \showarticletitle{Bbn: Bilateral-branch network with cumulative
  learning for long-tailed visual recognition}. In
  \bibinfo{booktitle}{\emph{CVPR}}.
\newblock


\bibitem[Zhu et~al\mbox{.}(2022)]%
        {zhu2022balanced}
\bibfield{author}{\bibinfo{person}{Jianggang Zhu}, \bibinfo{person}{Zheng
  Wang}, \bibinfo{person}{Jingjing Chen}, \bibinfo{person}{Yi-Ping~Phoebe
  Chen}, {and} \bibinfo{person}{Yu-Gang Jiang}.}
  \bibinfo{year}{2022}\natexlab{}.
\newblock \showarticletitle{Balanced contrastive learning for long-tailed
  visual recognition}. In \bibinfo{booktitle}{\emph{CVPR}}.
\newblock


\end{thebibliography}

\newpage
\appendix

\section{Implementation Details}
\label{sec:sup_implementation_details}
Before introducing the specific setting of different hyperparameters in our experiments, we first present the symbols and their corresponding definitions in~\cref{tab:parameter-list}. Throughout all experiments, we set the value of $\lambda_u$=1, while the other parameters' specific settings will be explained below. Additionally, at the initial stage of training, we perform model warmup by disregarding the unlabeled data in the loss calculation because the pseudo labels are unreliable. Specifically, we carry out 10 epochs of warmup for ImageNet-LT and 20 epochs for other datasets.

\fakeparagraph{ImageNet-LT.}
For both the 20$\%$ and 50$\%$ labeled subsets, we set $\tau$ to 0.7, $\beta$ to 3 and $\lambda_m$ to 0.75. Regarding the 20$\%$ labeled subset, we set $\alpha$ to 0.5, whereas for the 50$\%$ labeled subset, we set it to 0.75. Similarly, we set $\lambda$ to 1.25 for the 20$\%$ labeled subset and 0.75 for the 50$\%$ labeled subset.

\fakeparagraph{ImageNet127.}
We assess \system using the 1$\%$ and 10$\%$ labeled subsets of ImageNet127, with resolutions of 32$\times$32 and 64$\times$64. We maintain the value of $\tau$ to 0.95 for all experiments, while the other parameter settings for each setup are presented in~\cref{tab:imagenet127_parameters}.

\fakeparagraph{CIFAR-LT.}
We set $\tau$=0.95, $\alpha$=1.5 and $\beta$=3 for both CIFAR10-LT and CIFAR100-LT. Additionally, we set $\lambda$=0.75 and $\lambda_m$=0.25 for CIFAR10-LT, and $\lambda$=1.25 and $\lambda_m$=1.25 for CIFAR100-LT.

\fakeparagraph{Mismatched ImageNet.}
When conducting experiments on ImageNet with mismatched labeled and unlabeled sets, including $\gamma_u$=1, 20 and 100, we set the following hyperparameters: $\tau$=0.7, $\alpha$=1, $\beta$=3, $\lambda$=0.5, and $\lambda_u$=0.75.

\fakeparagraph{Ablation studies.}
We carry out ablation studies on the CIFAR100-LT dataset with $\gamma=10$. For these experiments, we maintain consistency with the main experiments except for the specific parameter being explored.

\begin{table}[htbp]
\centering
\resizebox{\columnwidth}{!}{%
\renewcommand\arraystretch{1.2}
\begin{tabular}{cl}
\toprule
\multicolumn{1}{c}{symbol} & \multicolumn{1}{c}{meaning}  \\ \cmidrule(lr){1-1} \cmidrule(lr){2-2}
$\tau$       &  the threshold above which we retain a pseudo label         \\
$\alpha$     &  controls the variance of weights in adaptive weighting     \\
$\beta$      &  controls the re-balancing degree of the memory bank     \\
$\lambda$    &  controls the degree of reversed sampling \\
$\lambda_u$  &  the relative weight of the loss term $\mathcal{L}_{u}$      \\
$\lambda_m$  &  the relative weight of the loss term $\mathcal{L}_{mem}$    \\
\bottomrule
\end{tabular}%
}
\smallskip
\caption{A list of hyperparameters and their respective definitions.}
\label{tab:parameter-list}
\end{table}

\begin{table}[htbp]
\centering
\resizebox{0.7\columnwidth}{!}{%
\begin{tabular}{cccccc}
\toprule
labeled ratio & resolution & $\alpha$ & $\beta$ & $\lambda$ & $\lambda_m$ \\
\cmidrule(lr){1-2} \cmidrule(lr){3-6}
1$\%$   &  32$\times$32  &  2    &  1  &  0.75  & 0.5  \\
1$\%$   &  64$\times$64  &  1.75 &  1  &  1     & 0.5  \\  
10$\%$  &  32$\times$32  &  1.5  &  1  &  1     & 0.75 \\  
10$\%$  &  64$\times$64  &  1.5  &  1  &  1.25  & 1    \\  
\bottomrule
\end{tabular}%
}
\caption{Hyperparameter settings for ImageNet127 dataset.}
\label{tab:imagenet127_parameters}
\end{table}

\section{More Experimental Results}
\label{sec:sup_more_experiment_results}
We conduct additional ablation experiments in this section to gain further insight into \system.

\begin{figure}[htbp]
  \centering
  \includegraphics[width=0.8\linewidth]{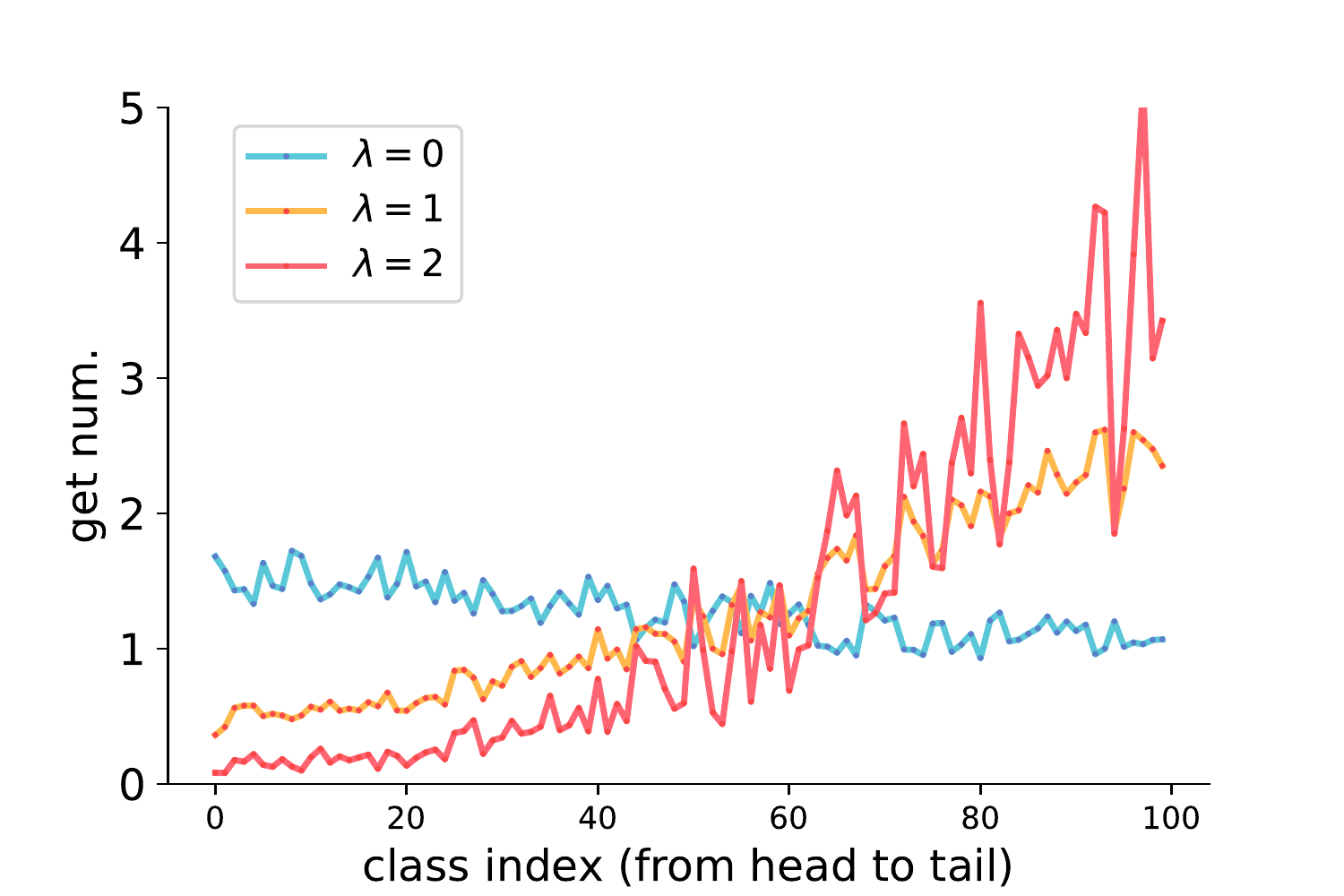}
  \caption{The distribution of data sampled from the memory bank, and a larger $\lambda$ leads to a more reversed result.}
  \label{fig:get_distribution}
\end{figure}

\begin{figure}[htbp]
  \centering
  \includegraphics[width=0.75\linewidth]{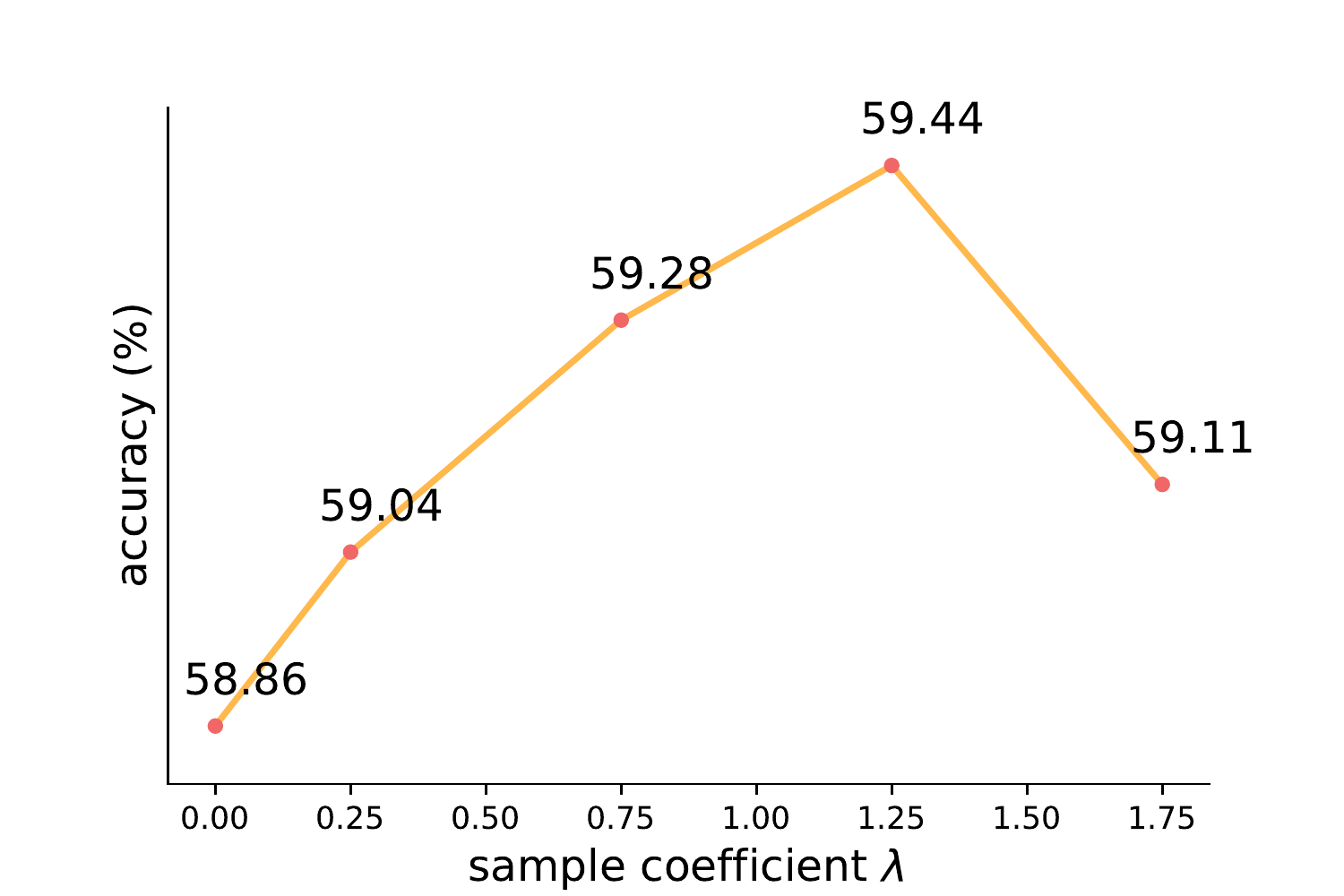}
  \caption{The accuracy achieved with varying values of $\lambda$.}
  \label{fig:sp_acc}
\end{figure}

\fakeparagraph{Degree of reversed sampling from the memory bank.}
When re-sampling from the memory bank, we adopt a reversed sampling strategy, where the parameter $\lambda$ controls the extent of reversal. A larger value results in a higher probability of selecting minority classes, and the experimental results visualized in~\cref{fig:get_distribution} are consistent with this. We plot the accuracy achieve with different values of $\lambda$ in~\cref{fig:sp_acc} and observe that a moderate value is suitable, as excessively large or small will lead to deteriorate results.

\begin{table}[htbp]
\centering
\resizebox{0.48\columnwidth}{!}{%
\renewcommand\arraystretch{1.2}
\begin{tabular}{ccc}
\toprule
weak & strong & accuracy (\%) \\ \cmidrule(lr){1-2} \cmidrule(lr){3-3}
\Checkmark   &                &   59.0    \\
             &  \Checkmark    &   \textbf{59.4}    \\
\Checkmark   &  \Checkmark    &   58.7    \\  
\bottomrule
\end{tabular}%
}
\smallskip
\caption{The model's accuracy when different features are stored in the memory bank.}
\label{tab:exp_ablate_features_self}
\end{table}

\fakeparagraph{Different configurations of the memory bank.}
The \system employs a memory bank to cache the features of the unlabeled data along with their pseudo labels. By default, only the strongly augmented feature $E(\mathcal{A}(u_j)$ is in the memory bank, where $E(\cdot)$ denotes the feature extractor. However, each unlabeled sample undergoes two different augmentations, which produces two different versions of features, namely $E(\mathcal{A}(u_j))$ and $E(\alpha(u_j))$. Consequently, there are multiple configurations of the memory bank, and we can store only one version or both of them. As shown in~\cref{tab:exp_ablate_features_self}, the model performs well in all cases, and achieves the best result when only $E(\mathcal{A}(u_j))$ is stored.

\end{document}